%% file: acl_latex.tex
\pdfoutput=1

\documentclass[11pt]{article}

\usepackage{acl}

\usepackage{times}
\usepackage{latexsym}

\usepackage[T1]{fontenc}

\usepackage[utf8]{inputenc}

\usepackage{microtype}

\usepackage{inconsolata}

\usepackage{amsmath}
\usepackage{pifont}
\usepackage{booktabs}
\usepackage{blindtext}
\usepackage[para,online,flushleft]{threeparttable}
\usepackage{algorithm}
\usepackage{algpseudocode}
\usepackage{graphicx}
\usepackage{subfigure} 
\usepackage{enumitem}
\usepackage{multirow}
\usepackage{arydshln}
\usepackage{verbatim}
\usepackage{sourcecodepro}
\usepackage{tcolorbox}
\tcbuselibrary{raster}
\tcbuselibrary{breakable}
\usepackage{listings}
\usepackage{soul}
\usepackage{makecell}
\usepackage{colortbl}
\usepackage{xcolor}
\usepackage{multicol}  
\usepackage[figuresright]{rotating}
\usepackage{amssymb}
\usepackage{pifont}
\usepackage{framed}

\input{macros}

\title{\model{}: Tool-augmented Language Models for Scientific Reasoning}

\author{
 Yubo Ma$^{1\ast}$, Zhibin Gou$^{2\ast}$, Junheng Hao$^{3}$, Ruochen Xu$^{3}$, Shuohang Wang$^{3}$, \\ \textbf{Liangming Pan$^4$, Yujiu Yang$^2$, Yixin Cao$^5$ , Aixin Sun$^1$, Hany Awadalla$^{3}$, Weizhu Chen$^{3}$} \\
 $^1$ Nanyang Technological University
 $^2$ Tsinghua University
 $^3$ Microsoft \\
 $^4$ University of California, Santa Barbara
 $^5$ Singapore Management University\\
\texttt{yubo001@e.ntu.edu.sg}\\
}

\begin{document}
\maketitle
\renewcommand{\thefootnote}{\fnsymbol{footnote}}
\footnotetext[1]{This work is done during Yubo and Zhibin’s internship at Microsoft.}
\renewcommand{\thefootnote}{\arabic{footnote}}

\input{abstract}
\input{1_intro}
\input{2_task}
\input{3_training-dataset}
\input{4_model}
\input{5_benchmark}
\input{6_experiments}
\input{7_conclusion}
\input{limitation}
\bibliography{custom}

\appendix
\label{sec:appendix}
\clearpage

\input{A_related-work}
\twocolumn
\input{B_training-details}
\input{C_eval-details}
\input{D_annotation-details}
\onecolumn
\input{E_appendix-case}
\input{F_prompt}

\end{document}

%% file: macros.tex
\definecolor{darkgreen}{RGB}{50,100,0}
\definecolor{darkred}{RGB}{200, 0, 0}
\definecolor{lightred}{RGB}{250, 200, 200}
\definecolor{lightblue}{RGB}{200, 200, 250}
\newcommand{\blue}{\cellcolor{lightblue}}

\definecolor{color_solution}{RGB}{255, 126, 121}
\definecolor{color_tool}{RGB}{91, 155, 213}
\newcommand{\cmark}{\textcolor{darkgreen}{\ding{51}}} 
\newcommand{\xmark}{\textcolor{darkred}{\ding{55}}} 

\newcommand{\ie}{\emph{i.e.,}\xspace}
\newcommand{\eg}{\emph{e.g.,}\xspace}
\newcommand{\etc}{\emph{etc}\xspace}

\newcommand{\model}[1]{\textsc{SciAgent}} 
\newcommand{\corpus}[1]{\textsc{MathFunc}} 
\newcommand{\bench}[1]{\textsc{SciToolBench}} 

\lstdefinestyle{python}{
    language=Python,
    basicstyle=\fontsize{7}{9.5}\ttfamily,
    keywordstyle=\color{blue},
    commentstyle=\color{gray},
    stringstyle=\color{black},
    showstringspaces=false,
    breaklines=true,
    breakindent=0pt,
    breakatwhitespace=false,
    escapeinside={(*@}{@*)}
}

\lstdefinestyle{plain}{
    basicstyle=\fontsize{7}{9.5}\ttfamily,
    keywordstyle=\color{blue},
    commentstyle=\color{gray},
    stringstyle=\color{green},
    showstringspaces=false,
    breaklines=true,
    breakatwhitespace=false,
    breakindent=0pt,
    escapeinside={(*@}{@*)}
}

\lstdefinestyle{prompt}{
  basicstyle=\footnotesize\ttfamily,
  columns=fullflexible,
  breaklines=true,
  frame=none,
  extendedchars=true,
  escapechar=@,
  literate={á}{{\'a}}1 {ã}{{\~a}}1 {é}{{\'e}}1 {£}{{\pounds}}1 {–}{{-}}1 {’}{{'}}1,
  frame=lines
}

%% file: abstract.tex
\begin{abstract}
Scientific reasoning poses an excessive challenge for even the most advanced Large Language Models (LLMs). To make this task more practical and solvable for LLMs, we introduce a new task setting named \textit{tool-augmented scientific reasoning}. This setting supplements LLMs with scalable toolsets, and shifts the focus from pursuing an omniscient problem solver to a proficient tool-user. To facilitate the research of such setting, we construct a tool-augmented training corpus named \corpus{} which encompasses over 30,000 samples and roughly 6,000 tools. Building on \corpus{}, we develop \model{} to retrieve, understand and, if necessary, use tools for scientific problem solving. Additionally, we craft a benchmark, \bench{}, spanning five scientific domains to evaluate LLMs' abilities with tool assistance. Extensive experiments on \bench{} confirm the effectiveness of \model{}. Notably, \textsc{SciAgent-Mistral-7B} surpasses other LLMs with the same size by more than 13\% in absolute accuracy. Furthermore, \textsc{SciAgent-DeepMath-7B} shows much superior performance than ChatGPT.
\end{abstract}

%% file: 1_intro.tex
\section{Introduction}

\textbf{\textit{Scientific reasoning}}~\cite{ouyang2023structured, zhao2023knowledgemath} aims to comprehend and make decisions regarding problems among STEM (\textit{Science, Technology, Engineering and Mathematics}) domains. It is a fundamental aspect of intelligence, a demanding capability of Large Language Models (LLMs), and a notoriously challenging task. For instance, even GPT-4~\cite{openai2023gpt4} achieves only $50\%$ and $35\%$ accuracy on TheoremQA~\cite{chen-etal-2023-theoremqa} and SciBench~\cite{wang2023scibench}, respectively. Regarding open-source LLMs such as LLaMA-2~\cite{touvron2023llama} and CodeLlama~\cite{rozière2023code}, their performances are only about $10\%$ accuracy or even less.

\begin{figure}[!t]
\centering
    \includegraphics[width=0.9\linewidth]{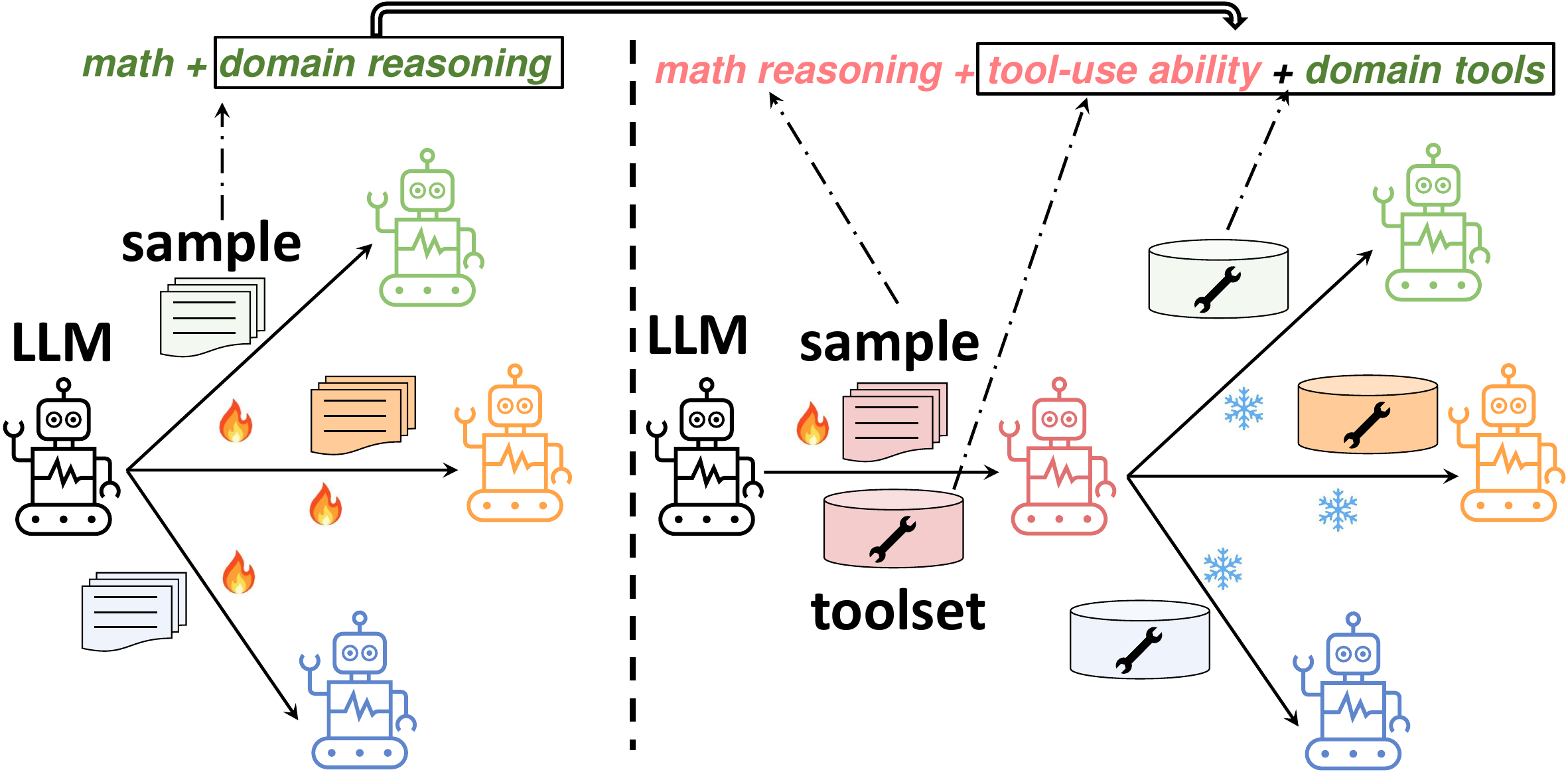}
    \caption{Two paradigms for scientific reasoning. Different colors represent different scientific domains. \textbf{Left:} Collecting annotations and fine-tuning LLMs domain by domain.
    \textbf{Right:} Our proposed \textit{tool-augmented} setting. LLMs are fine-tuned on math-related, tool-augmented samples (color in red). When adapting LLMs to a specific domain, a pluggable and domain-specific toolset is attached. No additional fine-tuning is further required.
    }
    \label{figs: intro}
\end{figure}

The challenge in scientific reasoning arises from the need for both mathematical (math) and domain-specific reasoning abilities. To address the physical problem in Figure~\ref{figs: pipeline}, for example, it is necessary to both understand \textit{Malus' law} (domain knowledge) for analyzing the intensity of polarized light, and possess quantitative ability for calculating the light intensity ratios. A natural approach involves collecting annotations and fine-tuning LLMs to enhance their math and domain-specific reasoning abilities, as depicted in Figure~\ref{figs: intro} (left). 
However, annotating scientific reasoning problems is extremely expensive. What is worse, adapting LLMs to a new domain demands a fresh round of annotation and fine-tuning, rendering this approach impractical.

In this paper, we draw inspirations from \textit{tool learning}~\cite{qin2023tool} to enhance LLMs' scientific reasoning capabilities.
Instead of solving scientific problem from scratch, humans have summarized and wrapped various points as generalized and well-documented functions in scientific computing softwares, such as Matlab, WolframAlpha, SymPy, \etc. These functions\footnote{In this work, tools refer to Python functions. We use tools and functions interchangeably unless otherwise specified.}, which could be equivalently viewed as external tools, greatly facilitate math-adept users to solve difficult scientific problems. In analogy with humans, we do not pursue an omniscient \textbf{solver} across various scientific domains. Instead, we assume the access to domain-specific toolsets and purse a unified, generalized LLM-based \textbf{tool-user} as shown in the Figure~\ref{figs: intro} (right). 
This approach tackles domain-specific reasoning challenges by enabling LLMs learn to use a reusable and scalable toolkit. It alleviates the reasoning challenges of LLMs by concentrating solely on enhancing their tool-use abilities.
These abilities are not only easier to acquire but also applicable across a variety of scientific fields. By attaching domain-specific toolsets, our tool-users can be readily adapted to different fields without the need for additional in-domain fine-tuning.

This work focuses on developing and benchmarking the ability of LLMs in scientific reasoning \textbf{with the help of tools}.  
We envision a scenario where LLMs have access to a domain-specific toolset, comprising various specialized functions. Upon this scenario, we propose a complete framework of dataset construction, model training and evaluation.
Given a scientific question, LLMs are supposed to retrieve functions from the toolset and optionally incorporate functions into the formulated solution. We employ an automatic pipeline featuring GPT-4 to compile a large-scale, math-related, tool-augmented training corpus named as \corpus{}. This corpus is designed to enable LLMs to learn both essential math skills and how to retrieve, understand and use functions properly. 
As a result, \corpus{} contains 31,375 samples and equipped with a toolset encompassing 5,981 generalized and well-documented functions. We detail this training corpus in Section~\ref{sec:training corpus}.

We fine-tune open-source LLMs on \corpus{} to develop tool-augmented agents named \model{} detailed in Section~\ref{sec:model}. As shown in Figure~\ref{figs: pipeline}, \model{} firstly generate a \textit{high-level planning} in response to a given question. The agents then use this plan, along with the question, to retrieve functions from the given toolset. Leveraging these retrieved functions, the agents further complete the \textit{low-level action} integrating natural language and Python code. Finally the agents execute the code to complete the problem at hand.

To benchmark the tool-use abilities in scientific reasoning, we develop a new benchmark named \bench{} as described in Section~\ref{sec: benchmark}. Building upon TheoremQA~\cite{chen-etal-2023-theoremqa} and SciBench~\cite{wang2023scibench}, it has 856 questions covering five domains: \textit{Mathematics}, \textit{Physical}, \textit{Chemistry}, \textit{EECS}, and \textit{Finance}. It also contains five domain-specific toolsets comprising a total of 2,446 functions.
We evaluate \model{} on \bench{} and another benchmark derived from CREATOR-challenge~\cite{qian-etal-2023-creator}. Experimental results demonstrate that our agents present remarkable scientific reasoning capabilities. Notably, \textsc{SciAgent-Mistral-7B} surpasses the best comparable open-source LLMs by an absolute 13.4\% accuracy, and \textsc{SciAgent-DeepMath-7B} outperforms ChatGPT by a large margin. We also conduct an extensive analysis of the benefits and limitations of \model{} series, providing valuable insights for future research.

%% file: 2_task.tex
\section{Preliminary}

\noindent \textbf{Related Work.} 
Current methods~\cite{chen-etal-2023-theoremqa, xu2023lemur, ouyang2023structured}, especially those based on open-source LLMs, perform far from satisfactory on scientific reasoning benchmarks~\cite{chen-etal-2023-theoremqa, wang2023scibench}. We attribute it to the scarcity of annotated samples across diverse scientific domains. As a comparison, LLMs present much more remarkable performance on math problems~\cite{yue2023mammoth, gou2023tora, azerbayev2023llemma} due to the abundant training corpora and/or annotations. Different from concurrent work~\cite{zhang2024sciglm} which collects physics and chemistry annotations, we do not pursue a problem-solver on some specific scientific domains. Instead, we consider to develop a generalized tool-user being proficient on solving diverse scientific problems with the aid of tools. Following previous work on math domain~\cite{qian-etal-2023-creator, cai2023large, yuan2023craft}, the tools here refer to Python functions. Please see more detailed literature review in Appendix~\ref{appendix:related_work}.

\noindent \textbf{Task Formulation.}
Given a scientific domain $D$ (\eg physics), \textit{tool-augmented} scientific reasoning task assumes access to (1) a question $q \in D$ and (2) a toolset $F_D$. $F_D$ encompasses large amounts of well-documented, domain-specific functions $\{f_1, ..., f_m\}$. Our objective is to develop an agent $\mathcal{M}$ which selectively use functions in $F_D$ to enhance the answering for the question $q$. 

%% file: 3_training-dataset.tex
\begin{figure*}[!t]
\centering
    \includegraphics[width=\textwidth]{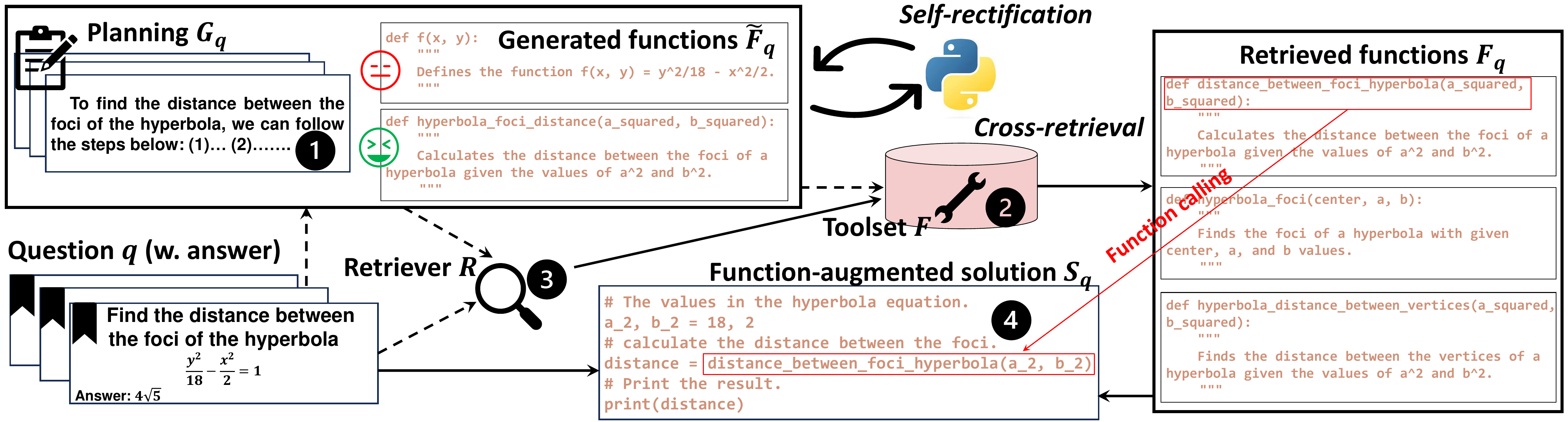}
    \caption{Automatic pipeline for \corpus{} construction. Please view it starting from the bottom left corner and proceed clockwise. We disentangle the constructions of toolset (dashed lines) and function-augmented samples (solid lines) for more generalized annotations. We do not visualize the function-free samples for simplicity.}
    \label{figs: training_annotation}
\end{figure*}

\section{Training Corpus: \corpus{}}
\label{sec:training corpus}
To our best knowledge, there are no readily available tool-augmented datasets in scientific reasoning domains. Therefore, we construct a corpus named \corpus{} teaching LLMs to better understand and use functions. \corpus{} is composed of (1) a toolset $F$\footnote{We remove the domain-specific subscript $D$ for expression simplicity. The same below.} including 5,981 generalized, well-documented, math-related functions and (2) a dataset $D$ encompassing 31,375 samples in which solutions call the function from the toolset if necessary (\eg \textcircled{4} in Figure~\ref{figs: training_annotation}). We build this corpus based on MATH~\cite{hendrycksmath2021} training set because we expect to teach LLMs both math skills and tool-use abilities. 

\noindent \textbf{Sample Format.}
Each sample is a quintuple $(q, G_q, F_q, S_q, a_q)$. Here $q$ is a question, $G_q$ is the planning, $F_q$ is the function set filtered from the toolset ($F_q \subset F$, $|F_q| \ll |F|$), $S_q$ is the solution and $a_q$ is the answer. $S_q$ interleaves rationales $E_q$\footnote{Here $E_q$ is written in natural language but formatted as the annotation lines in the program.} and programs $P_q$ which optionally call functions in $F_q$ to facilitate the problem solving.

We employ an automatic pipeline to construct \corpus{}. We illustrate the pipeline in Figure~\ref{figs: training_annotation} and detail the process in the following subsections.

\subsection{Planning and Toolset Construction}
\label{subsec:Planning and Toolset Construction}
This module is depicted in the top-left side of Figure~\ref{figs: training_annotation}. Given a question $q$ and its ground-truth solution (written in pure natural language) in MATH training set, we ask GPT-4 to generate (1) a high-level planning $G_q$ to analyze this question, (2) one or more well-documented functions $\tilde{F}_q$ and (3) a solution $\tilde{S}_q$ calling the functions above. The prompt used is shown in Appendix~\ref{appendix:function_collection}. In the prompt, we emphasize that the functions should be as \textbf{composable and generalized} as possible. Specifically, we do not hope that each question generates only one ad-hoc function (which could only be used by this question). Instead, we expect GPT-4 to generate functions that follow the points in the planning $G_q$ and can be reused by other questions. Following previous work~\cite{qian-etal-2023-creator, pan-etal-2023-logic}, we provide the error feedback to GPT-4 if the solutions fail to execute, and ask GPT-4 to rectify the errors in $\tilde{F}_q$ or $\tilde{S}_q$. We repeat this procedure until successful execution or reaching maximum loop limitation. The prompt used for rectification is shown in Appendix~\ref{appendix:self_rectification}.

We collect $G_q$ (\textcircled{1} in Figure~\ref{figs: training_annotation}, the same below) and add $\tilde{F}_q$ to the toolset (\textcircled{2}) for question $q$ if the rectified solution $\tilde{S}_q$ leads to the correct answer $\tilde{a}_q$. Regarding the toolset, it is iterated on all questions and finally accumulated as below:

\begin{align*}
    F = \bigcup_{q \in D} \tilde{F}_q \cdot \text{I}(\tilde{a}_q \text{ is correct})
\end{align*}

\begin{figure*}[!htbp]
\centering
    \includegraphics[width=0.95\textwidth]{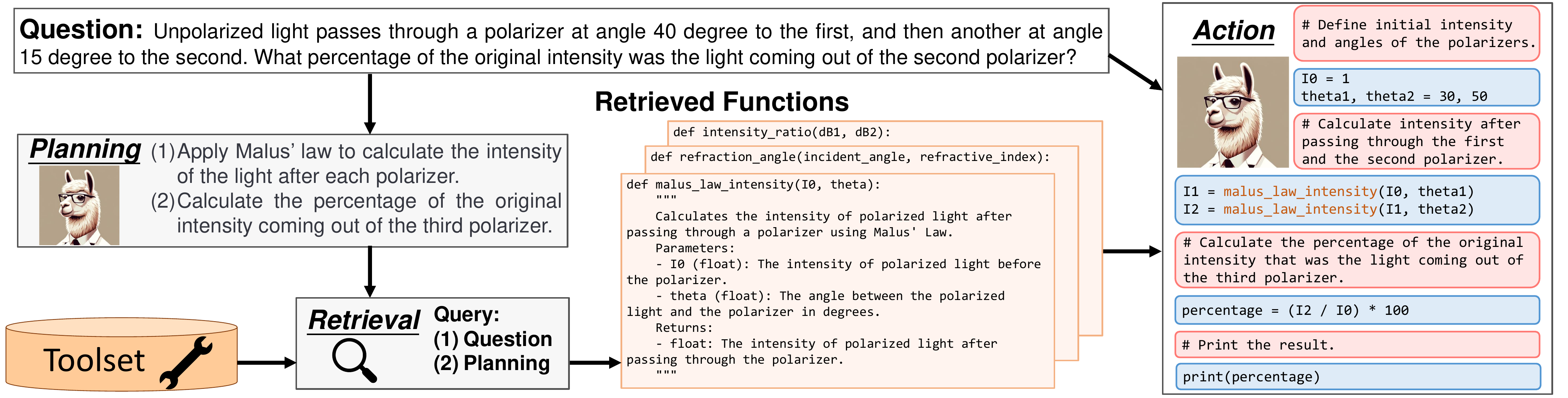}
    \caption{The model architecture of \model{}. Given a domain-specific toolset \includegraphics[height=1em]{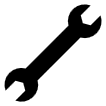}, our agent answers the question through four consecutive modules. (1) \textbf{Planning} \includegraphics[height=1em]{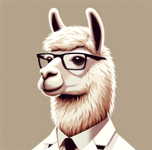}: provides a high-level plan for this problem. (2) \textbf{Retrieval} \includegraphics[height=1em]{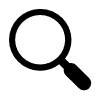}: retrieves related functions from attached toolset. (3) \textbf{Action} \includegraphics[height=1em]{figs/planner_actor.png}: generates a low-level solution interleaving rationale and program. The program uses the retrieved functions if necessary. (4) \textbf{Execution} \includegraphics[height=1em]{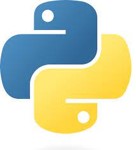}: calls Python executor to run the program and outputs the final answer. Not included in this figure for simplicity.}
    \label{figs: pipeline}
\end{figure*}

\subsection{Function-augmented Solutions}
\label{subsec: FAS}

To collect function-augmented solution $S_q$ and $F_q$, a natural idea is to directly use the $\tilde{S}_q$ and $\tilde{F}_q$ generated above. However, we find that $\tilde{S}_q$ tends to be contrived and specifically tailored to fit the requirements of function-calling. Moreover, some functions in $\tilde{F}_q$ tend to be ad-hoc\footnote{Despite we instruct GPT-4 to avoid generating ad-hoc functions, there are still some ad-hoc functions in $\tilde{F}_q$}. For examples, the function \texttt{f(x, y)} in Figure~\ref{figs: training_annotation} merely parameterizes the hyperbola for a specific question. Therefore we disentangle the construction of toolset and function-augmented solutions.
Given the developed toolset, we design a 
\textbf{cross-retrieval} strategy to retrieve more generalized functions $F_q$ and generate more qualified solutions $S_q$. Specifically, we remove $\tilde{F}_q$ from $F$ temporarily and then retrieve new functions $F_q \subseteq (F \backslash \tilde{F}_q)$ for question $q$. This strategy eliminates the likelihood of calling ad-hoc functions from $\tilde{F}_q$ in $S_q$. See examples of retrieved functions, all of which are derived from other questions, in the right side of Figure~\ref{figs: training_annotation}.

\noindent \textbf{Retriever.} 
The cross-retrieval strategy necessities a retriever because it is impractical to enumerate thousands of functions in $F \backslash \tilde{F}_q$. We train a dense retriever $R$ (\textcircled{3} in Figure~\ref{figs: training_annotation}). We concatenate the question $q$ and the generated planning $G_q$ as the query, and view the generated functions $\tilde{F}_q$ 
as the keys. See details about $R$ in Appendix~\ref{appendix:retriever_training}.

\noindent \textbf{Solution Generation.}
Upon the toolset $F$ and the retriever $R$, we retrieve three functions as $F_q$: 
\begin{align*}
    F_q = R([q, G_q]; F \backslash \tilde{F}_q)
\end{align*}

Then we employ GPT-4 to write solutions which optionally call functions in $F_q$ to generate the solution $S_q$ (\textcircled{4}). The prompt used is illustrated in Appendix~\ref{appendix:cross_retrieval}. We explicitly point out in the prompt that $f \in F_q$ should be called if and only if when they do lower the difficulty of problem solving. It mitigates the over-exploitation of function calling in $S_q$ and increases the robustness of models fine-tuned on these samples. Specifically, we firstly use GPT-4 with greedy decoding to generate solutions. For those failing to yield correct answers, we further apply nucleus sampling~\cite{holtzman2020curious} with 5 repeat times and 0.6 temperature. We filter wrong solutions and collect remaining 6,229 samples as our function-augmented solutions. 

In parallel, we use GPT-4 to generate function-free solutions. Though not indispensable, we expect them to further enhance the math reasoning, and accordingly the scientific reasoning, abilities of LLMs. We collect a total of 24,946 function-free solutions nucleus sampling with 5 repeat times and 0.6 temperature. These samples share similar format as ToRA-corpus~\cite{gou2023tora}, and do not retrieve/use any functions, \ie $F_q= \varnothing$.

%% file: 4_model.tex
\section{Model: \model{}}
\label{sec:model}
We develop \model{} for tool-augmented scientific reasoning task. It could make plan, retrieve functions, and leverage retrieved functions to facilitate the reasoning. We describe its inference procedure and training approach as below.

\subsection{Overview}
As shown in Figure~\ref{figs: pipeline}, \model{} comprises four successive modules.

\noindent \textbf{Planning.} This module provides a high-level profile for each question: $G_q = \mathcal{M}_\text{planning}(q)$. Such planning instructs a more targeted retrieval process.

\noindent \textbf{Retrieval.} Given the question and generated planning $G_q$, the retriever $\mathcal{M}_\text{retrieval}$ is introduced to retrieve related functions from the domain-specific toolset: $F_{q} = \mathcal{M}_\text{retrieval}([q, G_q]; F_D) \subseteq F_D$.

\noindent \textbf{Action.} This module aims to generate low-level solutions. Specifically, the agent produces $S_q = \mathcal{M}_\text{action}(q; F_{q})$. The solution $S_q$ is interleaved with natural language rationale $E_q$ and program snippet $P_q$. The program $P_q$ call retrieved functions with proper arguments if necessary. 

\noindent \textbf{Execution.} This module is simply a Python Executor to run the program $P_q$ for the final answer: $a_q = \text{Python-Executor}(P_q)$.

\subsection{Training}
Language models are used in three out of four modules in \model{}: planning, retrieval and action. 
Rearding retrieval, we directly use the retriever $R$ fine-tuned in Section~\ref{subsec: FAS} as $\mathcal{M}_\text{retrieval}$. For planning and action modules, they share the same LLMs: $\mathcal{M} = \mathcal{M}_\text{planning} = \mathcal{M}_\text{action}$. We fine-tune $\mathcal{M}$ with different instructions to make it act as planning and action modules, respectively. We construct instructions from $d=(q, G_q, F_q, S_q, a_q)$ in \corpus{}.

\begin{align*}
    D_\text{planning} &= \{(I_\text{plan}(q), G_q) | d \in D\} \\
    D_\text{action} &= \{(I_\text{action}(q, F_q), S_q) | d \in D\} 
\end{align*}

Here $I_\text{plan}$ and $I_\text{action}$ are instruction templates for planning and action modules. We show these instructions in Appendix~\ref{appendix:planning_and_action}, and mix up them as the training set $D = (D_\text{planning} \bigcup D_\text{action})$. Then we apply imitation learning on $D$ to fine-tune $\mathcal{M}$.

\begin{align*}
    L_\mathcal{M} = \sum_{(X, Y) \in D} - \text{log}\mathcal{P}(Y|X)
\end{align*}

\noindent \textbf{Implementation} We detail the training process of (1) the retriever $\mathcal{M}_\text{retrieval}$ and (2) the planner and actor $\mathcal{M}$ in Appendix~\ref{appendix:retriever_training} and~\ref{appendix:planning_and_action}, respectively.

%% file: 5_benchmark.tex
\section{Benchmark: \bench{}}
\label{sec: benchmark}

There currently exists no benchmark assessing the scientific reasoning capabilities of LLMs \textbf{when aided by tools}. To address this gap, we develop a benchmark called \bench{}. Our benchmark covers five domains: \textit{Mathematics (math)}\footnote{Our benchmark contains college-level questions on calculus, differential equations, group theory, \etc, which are different from the questions in our training corpus \corpus{}.}, \textit{Physics}, \textit{Chemistry}, \textit{Finance}, \textit{Electrical Engineering and Computer Science (EECS)}. Each domain is composed of a set of questions and a domain-specific toolset. The toolset consists of abundant generalized, high-quality and well-documented functions. We expect LLMs to retrieve, understand and, if necessary, use functions in it for reasoning.

\begin{table}[htbp!]
\small
\centering
\caption{The statistics of our benchmark. \textbf{\#Func}: Number of functions. \textbf{\#Pos./ \#Neg.}: The number of positive/negative functions in the toolset. \textbf{FPQ} (function per question): The number of derived positive functions from each question.}
\resizebox{\linewidth}{!}{
    \begin{tabular}{l|cccc}
    \toprule
    & \# Question & \# Func & \# Pos. / \# Neg. & Avg. FPQ \\
    \midrule
    \textbf{Math} & 434 & 1072 & 511 / 561 & 1.47 \\
    \textbf{Physics} & 156 & 534 & 243 / 291 & 1.63  \\
    \textbf{Chemistry} & 118 & 366 & 155 / 211 & 1.34 \\
    \textbf{Finance} & 66 & 253 & 97 / 156 & 1.62 \\
    \textbf{EECS} & 82 & 221 & 97 / 124 & 1.68 \\
    \midrule
    \textbf{All} & 856 & 2446 & 1103 / 1343 & 1.51 \\
    \bottomrule 
    \end{tabular}
    }
    \label{tbl:benchmark_stat}
\end{table}

\begin{figure}[htbp!]
\centerline{\includegraphics[width=0.9\linewidth]{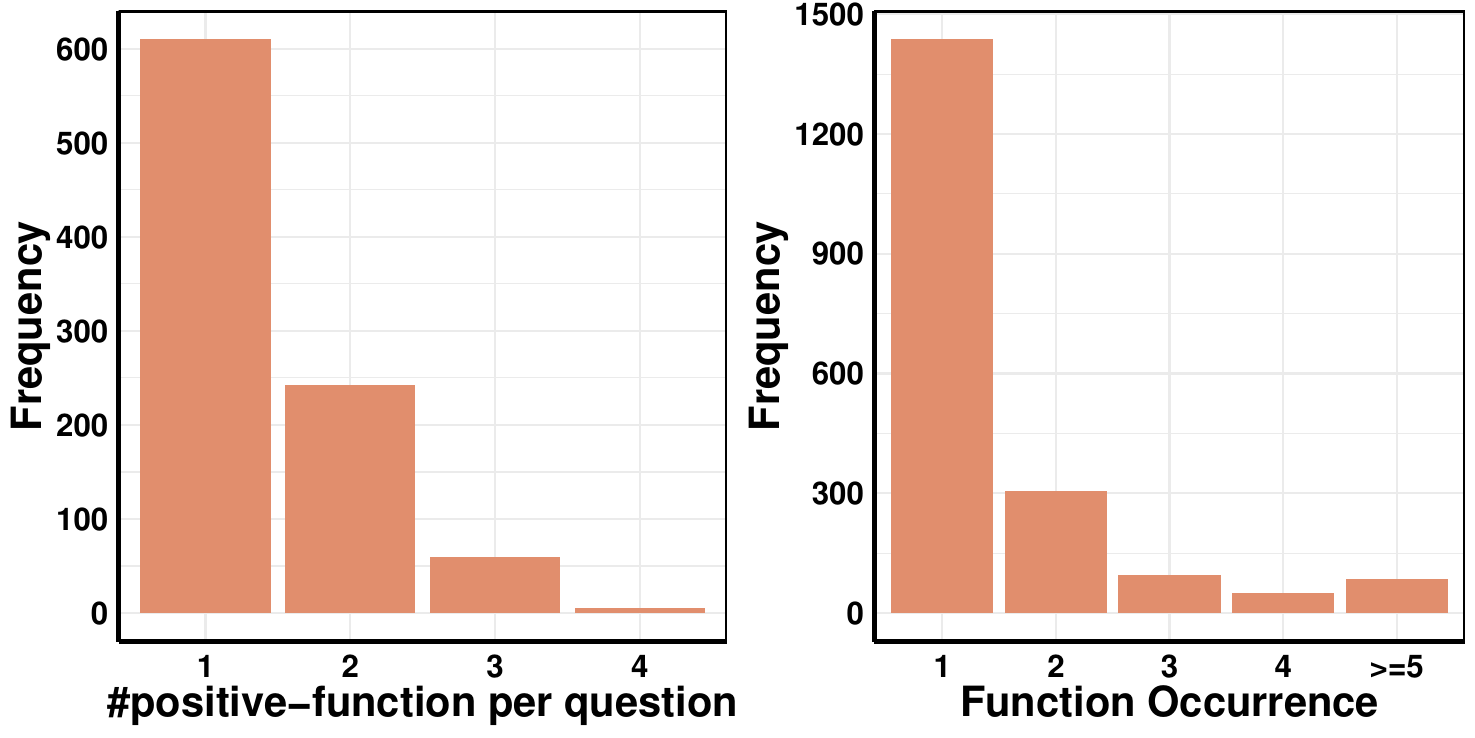}}
    \caption{\textbf{Left}: Histogram of FPQ (function per question). Higher values indicate greater composability. \textbf{Right}: Histogram of function occurrence. Higher values indicate more generalization and wider application.}
    \label{fig:benchmark_stat}
\end{figure}

\subsection{Dataset Overview.}
The statistics of \bench{} are presented in Table~\ref{tbl:benchmark_stat}. It comprises a total of 856 questions and 2,446 functions spanning across 5 scientific domains. Notably, \bench{} differs from previous tool-based benchmarks, such as Creation Challenge~\cite{qian-etal-2023-creator}, in several aspects: (1) Our benchmark encompasses a diverse range of scientific domains. (2) The tools provided are both composable and generalized across different questions. As indicated in Table~\ref{tbl:benchmark_stat}, each question requires an average of 1.51 functions for resolution. And as shown in Figure~\ref{fig:benchmark_stat}, over 500 functions are designed to be applicable to two or more questions, such as \texttt{integrate\_function} in math domain, \texttt{coulombs\_law} in physical domain, and \texttt{calculate\_pressure\_van\_der\_waals} in chemistry domain. It signifies that the functions in our toolset are not ad-hoc solutions tailored for specific questions. Instead, the effective utilization of the toolset demands significant reasoning abilities of tool-augmented LLMs. Thus we claim this benchmark challenging and practical.

\begin{figure}[!htbp]
\centering
    \includegraphics[width=\linewidth]{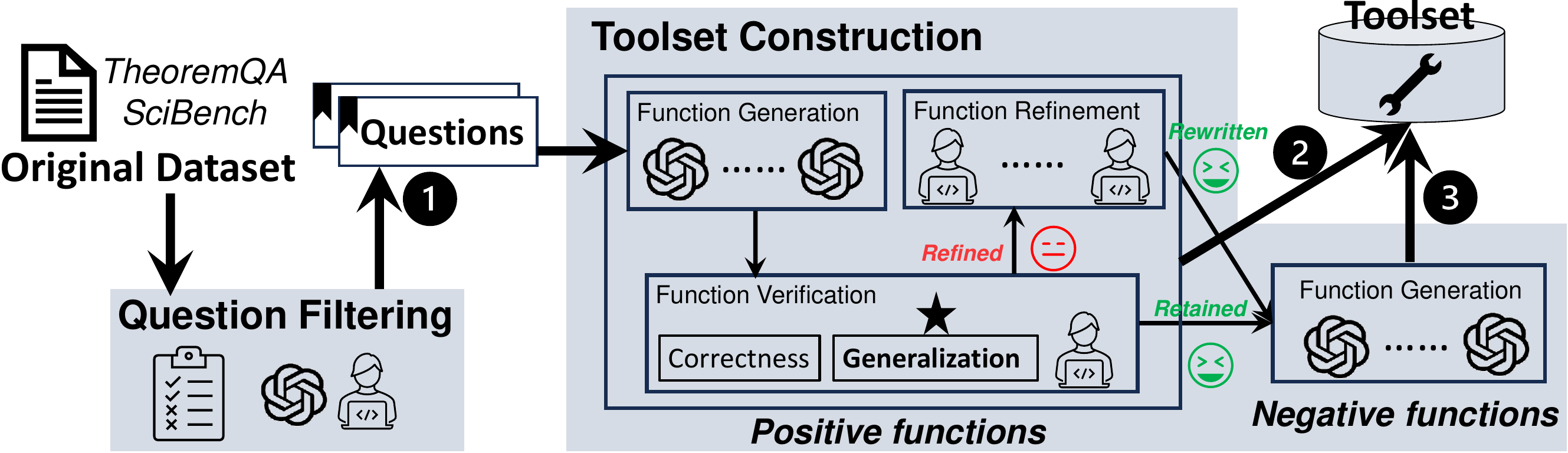}
    \caption{Semi-automatic annotation pipeline for \bench{}. \includegraphics[height=1em]{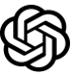}: GPT-4. \includegraphics[height=1em]{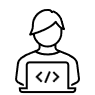}: Human annotator.}
    \label{figs: benchmark_annt}
\end{figure}

\begin{table*}[htbp!]
\centering
\caption{Main results on two benchmarks. We highlight our \model{} series in \colorbox{lightblue}{blue}.
The best results (among all open-source LLMs, the same below) are in bold face and the second best are underlined.
}
\small
    \begin{tabular}{lcc|c|cccccc}
    \toprule
        Model & Size & Toolset & \multirow{2}{*}{\textbf{CREATION}} & \multicolumn{6}{c}{\textbf{\bench{}}} \\ 
        & & & & Math & Physics & Chemistry & Finance & EECS & All \\
    \midrule
    \multirow{2}{*}{ChatGPT} & \multirow{2}{*}{-} & \xmark & 54.6 & 33.4 & 19.2 & 18.6 & 53.0 & 25.6 & 29.6 \\
    & & \cmark & 59.8 & 32.0 & 31.4 & 33.9 & 53.0 & 48.8 & 35.4 \\
    \multirow{2}{*}{GPT-4} & \multirow{2}{*}{-}  & \xmark & 60.0 & 52.8 & 42.9 & 47.5 & 65.2 & 35.4 & 49.5 \\
    & & \cmark & 69.8 & 63.1 & 63.5 & 63.6 & 80.3 & 80.5 & 66.2 \\ 
    \midrule
    LLaMA2 & 7B & \cmark & 12.6 & 4.3 & 10.9 & 8.4 & 13.6 & 11.0 & 8.3 \\ 
    CodeLlama & 7B & \xmark & 17.7 & 6.5 & 0.6 & 5.1 & 4.9 & 7.6 & 5.1\\
    CodeLlama & 7B & \cmark & 26.1 & 9.2 & 8.3 & 10.2 & 24.2 & 25.6 & 11.9 \\
    Llemma & 7B & \xmark & 26.4 & 10.4 & 4.5 & 8.5 & 10.6 & 7.3 & 8.8\\
    Llemma & 7B & \cmark & 34.3 & 16.4 & 21.2 & 14.4 & 36.4 & 22.0 & 19.1 \\
    Mistral & 7B & \xmark & 30.1 & 11.3 & 4.5 & 7.6 & 16.7 & 6.1 & 9.5 \\
    Mistral & 7B & \cmark & 27.6 & 13.1 & 13.5 & 14.4 & 34.8 & 19.5 & 15.6 \\
    Deepseek-Coder & 7B & \xmark & 36.8 & 20.3 & 8.3 & 5.9 & 22.7 & 12.2 & 15.5 \\
    Deepseek-Coder & 7B & \cmark & 31.3 & 21.0 & 15.4 & 10.2 & 30.3 & 36.6 & 20.7 \\
    Deepseek-Math & 7B & \xmark & 44.7 & 26.5 & 19.2 & 17.8 & 27.3 & 20.7 & 23.5 \\
    Deepseek-Math & 7B & \cmark & 41.3 & 24.2 & 24.4 & 25.4 & 43.9 & 42.7 & 27.7 \\
    ToRA-Coder & 7B & \xmark & 29.7 & 26.3 & 4.5 & 6.8 & 9.1 & 24.4 & 18.1 \\
    ToRA-Coder & 7B & \cmark & 21.4 & 21.7 & 4.5 & 5.1 & 13.6 & 15.9 & 15.1 \\
    MAmmoTH-Coder & 7B & \cmark & 21.6 & 14.8 & 18.5 & 11.0 & 25.8 & 40.0 & 19.7 \\
    \blue{\model{}\textsc{-Coder}} & \blue{7B} & \blue{\cmark} & \blue{53.0} & \blue{30.0} & \blue{28.3} & \blue{24.6} & \blue{39.3} & \blue{\underline{57.3}} & \blue{32.2}  \\
    \blue{\model{}\textsc{-Mistral}} & \blue{7B} & \blue{\cmark}& \blue{54.0} & \blue{31.3} & \blue{28.8} & \blue{22.9} & \blue{\underline{51.5}} & \blue{\textbf{61.0}} & \blue{34.1} \\
    \blue{\model{}\textsc{-DeepMath}} & \blue{7B} & \blue{\cmark} & \blue{\textbf{60.4}} & \blue{\textbf{41.2}} & \blue{\textbf{54.5}} & \blue{\textbf{44.9}} & \blue{\textbf{57.5}} & \blue{51.2} & \blue{\textbf{46.3}} \\
    \midrule
    LLaMA2 & 13B & \cmark & 23.3 & 12.2 & 11.5 & 6.8 & 22.7 & 14.6 & 12.4 \\ 
    CodeLlama & 13B & \xmark & 23.0 & 9.9 & 3.2 & 1.7 & 9.1 & 6.1 & 7.1 \\
    CodeLlama & 13B & \cmark & 38.9 & 12.7 & 14.7 & 7.6 & 33.3 & 34.1 & 16.0 \\
    ToRA-Coder & 13B & \xmark & 30.9 & 28.6 & 3.8 & 4.2 & 16.7 & 30.5 & 20.0 \\
    ToRA-Coder & 13B & \cmark & 28.0 & 32.0 & 2.6 & 11.9 & 24.2 & 35.4 & 23.6 \\
    MAmmoTH-Coder & 13B & \cmark & 34.7 & 21.4 & 18.6 & 11.0 & 25.8 & 39.0 & 21.5 \\
    \blue{\model{}\textsc{-Coder}} & \blue{13B} & \blue{\cmark} & \blue{\underline{54.4}} & \blue{\underline{35.0}} & \blue{\underline{32.1}} & \blue{\underline{28.8}} & \blue{42.4} & \blue{51.2} & \blue{\underline{35.7}} \\
    \bottomrule 
    \end{tabular}
    \label{tbl:main_results}
\end{table*}

\subsection{Dataset Annotation}
We design a pipeline shown in Figure~\ref{figs: benchmark_annt} to annotate the benchmark. It employs both GPT-4 and human annotators to combine their merits. We introduce it briefly as below and leave details in Appendix~\ref{appendix:details_of_scitoolbench}.

\noindent \textbf{Question Filtering}: We curate questions from TheoremQA~\cite{chen-etal-2023-theoremqa} and SciBench~\cite{wang2023scibench} to collect 856 questions (\textcircled{1} in Figure~\ref{figs: benchmark_annt}, the same below) in our benchmark. 

\noindent \textbf{Toolset Construction}: We construct domain-specific toolsets via two cascade modules: positive and negative function construction. We define positive functions (\textcircled{2}) as functions directly deriving from questions. The candidate positive functions (\textcircled{2}) are firstly generated from GPT-4. Then human annotators carefully check them and rewrite and/or remove the unqualified ones. We further automatically construct negative functions (\textcircled{3}) based on positive functions to reduce the shortcuts in our benchmark. We finally combine both positive and negative functions as the toolset in our benchmark.

%% file: 6_experiments.tex
\section{Experiments}
\subsection{Setup}
We conduct experiments on \bench{} to evaluate the tool-augmented scientific reasoning abilities of LLMs. We also employ CREATION Challenge~\cite{qian-etal-2023-creator} as the second benchmark. It comprises a total of 2,047 samples, with each sample consisting of a question and a ground-truth function. We aggregate all functions to assemble the toolset (thus including 2,047 functions). We report accuracy as the metric in all experiments.

\subsection{Baselines} We compare \model{} series with eight open-source LLMs: (1) LLaMA-2~\cite{touvron2023llama}, (2) CodeLlama~\cite{rozière2023code}, (3) Mistral~\cite{jiang2023mistral}, (4) Llemma~\cite{azerbayev2023llemma}, (5) Deepseek-Coder~\cite{guo2024deepseekcoder}, (6) Deepseek-Math~\cite{shao2024deepseekmath}, (7) MAmmoTH-Coder~\cite{yue2023mammoth} and (8) ToRA-Coder~\cite{gou2023tora}. 
We also list the performance of ChatGPT and GPT-4 for reference. We provide all LLMs the same retriever in Section~\ref{subsec: FAS} to retrieve functions from toolset (if attached). Please see more details in Appendix~\ref{appendix:eval_details}.

\begin{table*}[!htbp]
\centering
\small
\caption{Ablation study on \bench{}. We report the accuracy of samples across (1) all domains, (2) four domains excluding the math domain (wo. math). 
}
\resizebox{\linewidth}{!}{
    \begin{tabular}{l|cccc|cc|cc}
    \toprule
     & \textbf{Planning} & \textbf{Function-augmented} & \textbf{Function-free} & \textbf{Retriever} & \multicolumn{2}{c}{\textbf{Accuracy (7B)}} & \multicolumn{2}{c}{\textbf{Accuracy (13B)}} \\
     & & \textbf{solutions} & \textbf{solutions} & & All & wo. math & All & wo. math \\
    \midrule
    SciAgent-Coder & \cmark & \cmark (cross-retrieval) & \cmark & \cmark & \textbf{32.2} & \textbf{34.6} & \textbf{35.7} & \textbf{36.5} \\
    \midrule
    \multirow{3}{*}{\makecell{Intermediate variants \\ 1-3 }}  & \xmark & \cmark (cross-retrieval) & \cmark & \cmark & \underline{30.3} & \underline{33.9} & \underline{32.8} & \underline{34.4} \\
    & \xmark & \cmark (direct-use) & \cmark & \cmark & 17.8 & 17.3 & 26.6 & 31.0 \\
    & \xmark & \xmark & \cmark & \cmark & 26.3 & 26.1 & 30.4 & 31.7 \\
    \midrule
    CodeLlama & \xmark & \xmark & \xmark & \cmark & 11.9 & 14.7 & 16.0 & 19.4 \\
    \hspace{1em} \textit{wo. retriever} & \xmark & \xmark & \xmark & \xmark & 5.1 & 3.8  & 7.1 & 4.3  \\
    \bottomrule 
    \end{tabular}
}
\label{tbl:ablation_studies}
\end{table*}

\subsection{Main Results}
We fine-tune CodeLlama, Mistral and Deepseek-Math for yielding \model{}\textsc{-Coder}, \model{}\textsc{-Mistral} and \model{}\textsc{-DeepMath}. We show their results in Table~\ref{tbl:main_results} and observe: (1) Almost all LLMs present improved performance, \ie 5.3\% absolute and 61.6\% relative accuracy increase on average, when supplemented with toolsets. It validates the promise of the tool-augmented setting for scientific reasoning. (2) The models fine-tuned on math-related datasets from CodeLlama, \ie ToRA- and MAmmoTH-Coder, perform better than CodeLlama itself by 5.5\% abosolute accuracy. It presents the importance of essential math skills among diverse scientific domains. (3) Our agents consistently outperform other open-source LLMs by a large margin. Notably, \model{}\textsc{-Coder} surpasses the most competitive baseline, MAmmoTH-Coder, by absolute accuracy of 12.5\% and 14.2\% on the 7B and 13B versions. 
(4) Our strongest agent, \model{}\textsc{-DeepMath-7B}, substantially outperforms ChatGPT with toolset (46.3\% v.s. 35.4\%) and shows comparable results to GPT-4 without toolset (46.3\% v.s. 49.5\%). However, it still falls significantly behind GPT-4 when both are provided with the same tools. Such gap highlights the challenges of tool-augmented scientific reasoning (and our benchmark). 
(5) Both our agents and other baselines show relatively higher proficiency in the domains of math, finance, and EECS, but lower performance in physics and chemistry. We speculate that the first three domains align more closely with the training data's source distribution. Additional in-domain knowledge is demanding to further improve the performance in physics and chemistry domains.

\subsection{Ablation Study}
\label{subsec: ablation_study}
We investigate the effectiveness of components in our training data and agent modules. The specific variants we considered are as follows. (1) We remove the planning module in the agent. (2) We additionally drop the cross-retrieval strategy introduced in Section~\ref{subsec: FAS}.
In its place, we construct function-augmented solutions directly from $\tilde{F_q}$ and $\tilde{S_q}$. (3) We further remove all function-augmented solutions from our training data, and only keep the solutions without function callings (function-free solutions). (4) We do not fine-tune agents but merely use CodeLlama as $\mathcal{M}_\text{action}$ for inference. (5) We drop the retriever to disable the LLMs' tool-use abilities. Equivalently, it degrades to the baseline of CodeLlama + PoT~\cite{chen2022program} prompting.

We illustrate the performance of our agents and their ablated variants in Table~\ref{tbl:ablation_studies}. We observe that (1) Planning module significantly improves scientific reasoning abilities. As detailed and targeted queries for the retriever, the generated plannings increase the relatedness of retrieved functions. For instance, the function's Recall\texttt{@}3 increases from 48.3\% to 53.2\% in physics domain, and from 37.3\% to 39.8\% in chemistry domain. (2) The use of the cross-retrieval strategy is essential. Otherwise, the function-augmented solutions directly from $\tilde{F_q}$ and $\tilde{S_q}$ degrade the performance because they are too artificial and ad-hoc to teach LLMs using functions properly. (3) The absence of function-augmented solutions results in a performance drop (row 1 v.s. row 4 in Table~\ref{tbl:ablation_studies}) of 5.9\% and 5.3\% in absolute accuracy for 7B and 13B LLMs, respectively. It underscores the critical role of function-augmented solutions to enhance LLMs' tool-use abilities, and the necessity of our \corpus{} corpus. (4) The removal of function-free solutions (row 4 v.s. row 5) leads to an absolutely 14.4\% accuracy decrease. Specifically focusing on non-math samples, there is a notable performance drop of about 12\% as well. This clearly demonstrates the fundamental importance of math skills in diverse scientific reasoning tasks, and highlights how our math-related samples enhance LLMs' capabilities in this area. (5) Performance significantly declines when the retriever is removed. It illustrates that the retrieval module is crucial for accessing the appropriate functions from large-scale toolsets.  

\subsection{Analysis}
\noindent \textbf{Robustness of Toolsets.} We acknowledge the construction and maintenance of toolsets is sometime challenging. Therefore, we stress the importance of our agents' robustness. If a sub-par toolset were provided, an robust agent should at the very least perform comparably, if not better, than other competitive LLMs without tool-use. To evaluate the robustness of \model{}\textsc{-Coder}, we simulate two sub-par settings. (1) weak-related: for each question, we restrict the agents from retrieving functions that are directly derived from it. This setting greatly decreases the likelihood of retrieving a proper function from the toolset. (2) unrelated: we completely remove the domain-specific toolset in \bench{}. As a substitution, we provide the unrelated toolset constructed in \corpus{}.

\begin{table}[!htbp]
\centering
\small
\caption{Accuracy on \model{} with sub-par toolsets. \textbf{WR}: weak-related toolsets. \textbf{UR}: unrelated toolsets. \textbf{NA}: No toolset. The subscripts indicate the difference from the best LLMs (wo. toolsets) each column.}
\resizebox{\linewidth}{!}{
    \begin{tabular}{lc|cc|cc}
    \toprule
     \multirow{2}{*}{\textbf{Model}}  &  \multirow{2}{*}{\textbf{Toolset}} & \multicolumn{2}{c}{\textbf{Accuracy (7B)}} & \multicolumn{2}{c}{\textbf{Accuracy (13B)}} \\
     & & All & wo.math & All & wo. math \\
     \midrule
     \multirow{2}{*}{\makecell{SciAgent\\-Coder}} & WR & 18.8$_{\textcolor{green}{+0.7}}$ & 18.0$_{\textcolor{green}{+8.3}}$ &  24.6$_{\textcolor{green}{+4.6}}$ & 19.9$_{\textcolor{green}{+7.6}}$ \\
     & UR & 14.7$_{\textcolor{red}{-3.7}}$ & 10.7$_{\textcolor{green}{+1.0}}$ & 20.3$_{\textcolor{green}{+0.3}}$ & 14.7$_{\textcolor{green}{+2.4}}$ \\
    \midrule
    MAmmo-C & NA &  12.7 & 9.0 & 16.4 & 12.3 \\
    ToRA-C & NA & 18.1 & 9.7 & 20.0 & 11.1 \\
    \bottomrule 
    \end{tabular}
}
\label{tbl:robustness}
\end{table}

We compare our agents with two competitive LLMs, \ie ToRA-Coder and MAmmoTH-Coder, in above two settings. As shown in Table~\ref{tbl:robustness}, (1) \model{} series with unrelated toolsets present comparable performance with the two LLMs. In other words, our tool-augmented agents are unlikely to degrade the performance even under the extreme scenarios. (2) Our agents with weak-related toolsets significantly outperform the two LLMs, which further validates the robustness.

\begin{figure}[htbp!]
    \centering
    {\includegraphics[width=0.85\linewidth]{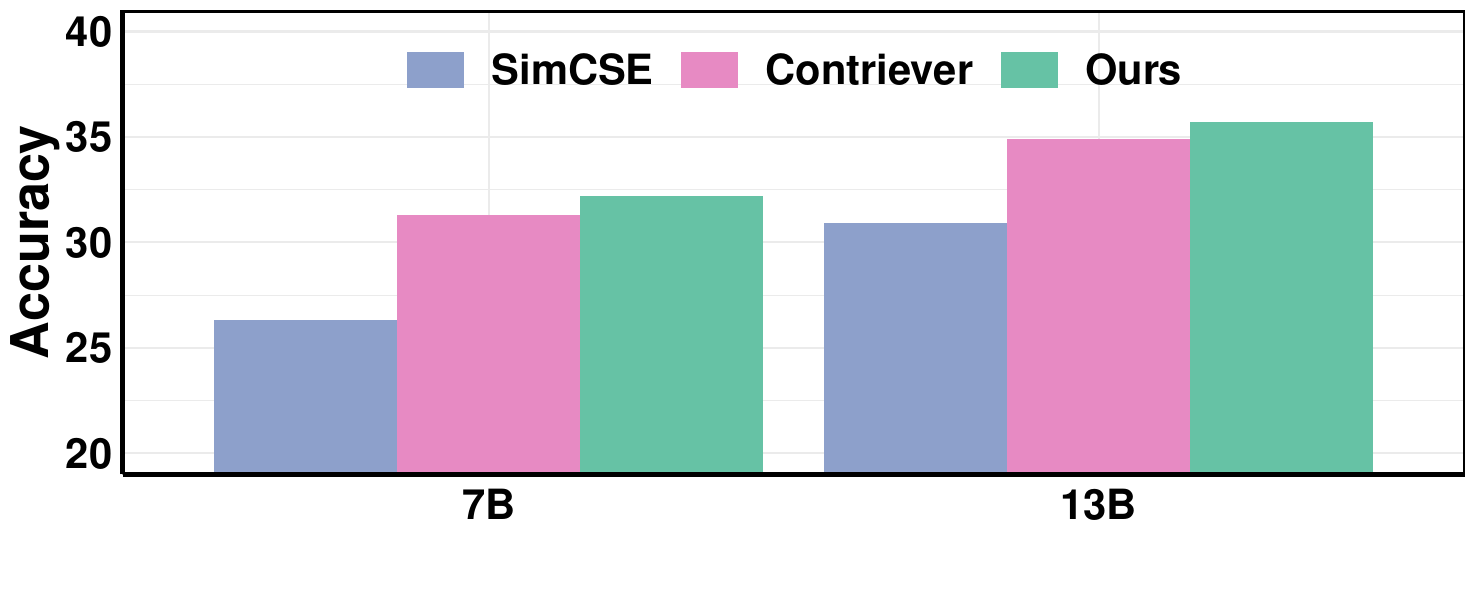}} 
    {\includegraphics[width=0.85\linewidth]{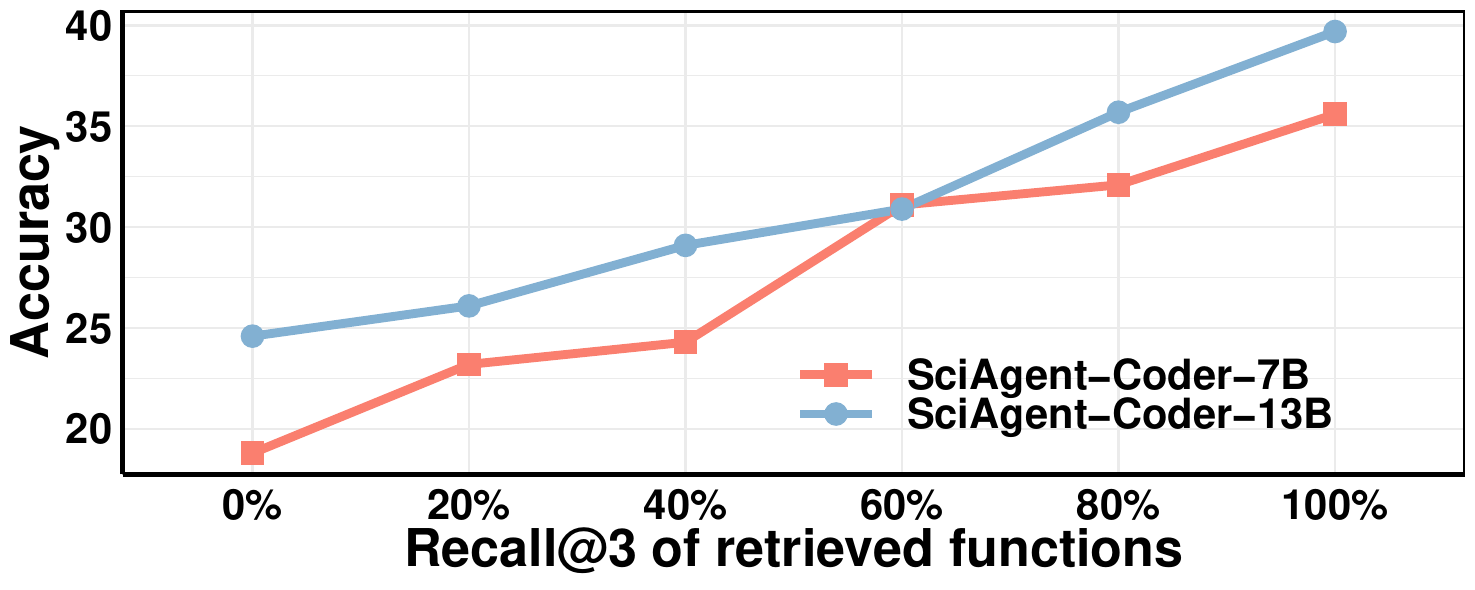}} 
    \caption{\textbf{Top}: Performance of \model{}\textsc{-Coder} on \bench{} with different retriever variants. \textbf{Bottom}: Relationship between the performance and the hit@3 of retrieved functions (artificially controlled).}
    \label{fig: retriever}
\end{figure}

\noindent \textbf{The Effect of Retriever Quality.} We explore the effect of retriever quality on the ending performance. We substitute our fine-tuned retriever in \model{} series by two competitive variants: SimCSE~\cite{gao-etal-2021-simcse} and Contriever~\cite{izacard2021contriever}. As shown in Figure~\ref{fig: retriever} (top), our retriever surpasses the other two. It shows that fine-tuning on the math domain benefits the retrieval of tools in the generalized scientific domains.

We further dive deep into the relationship between the hit ratio of tools and the agents' performance. To this end, we manually control the hit@3 ratio by artificially adding/removing the positive functions to/from the retrieved list. Results in Figure~\ref{fig: retriever} (bottom) show a clearly positive correlation between the hit ratio and the task accuracy. It illustrates that the retrieved functions facilitate the reasoning of scientific problems. However, we still observe a limit ($40\%$ accuracy) when the hit ratios reaching 100\%, showing the challenge of scientific reasoning even when aided by tools. We hope the future work to bridge this performance gap.

\begin{figure}[htbp!]
\centerline{\includegraphics[width=0.9\linewidth]{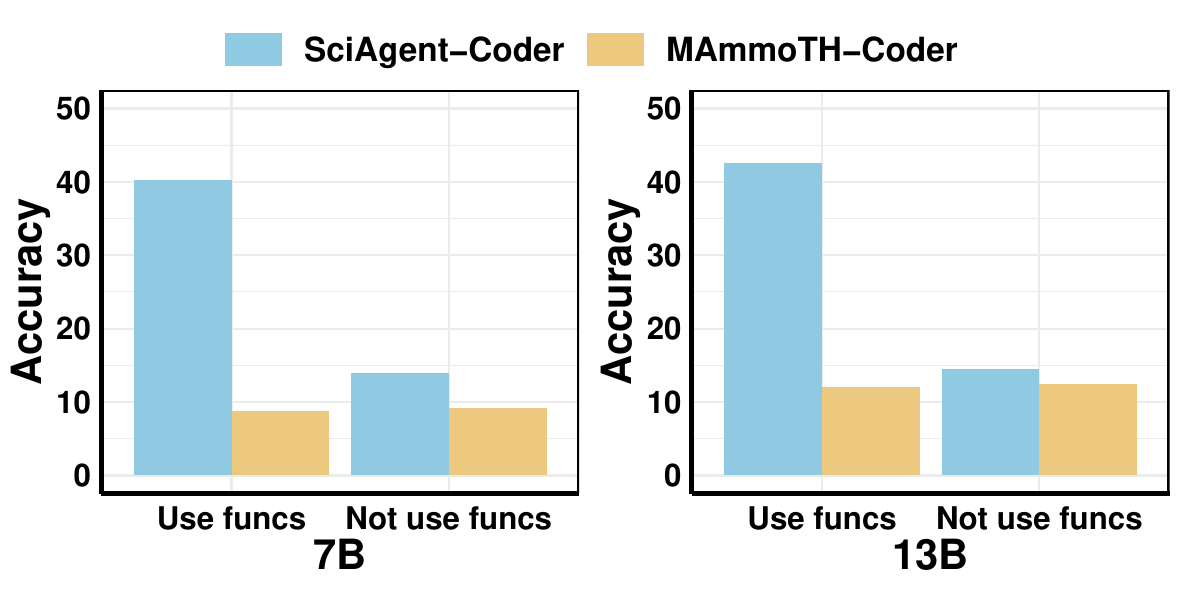}}
    \caption{The performance of \model{}\textsc{-Coder} (w. toolset) and MAmmoTH-Coder (wo. toolset) on samples which (1) use and (2) not use retrieved functions. 
    }
    \label{figs:analysis}
\end{figure}

\noindent \textbf{How the Retrieved Functions Benefit.}
To assess how the retrieved functions aid in the reasoning process of LLMs, we divided the samples into two subsets based on whether our agents use the retrieved functions to solve the problems. We evaluate the performance of these two subsets respectively, comparing with MAmmoTH-Coder series (without tool-use). The results in Figure~\ref{figs:analysis} reveal a two-fold benefit: (1) For samples where functions are explicitly called to solve the questions, our agents demonstrate a substantial 25\% improvement in absolute accuracy over LLMs that do not have access to functions. (2) Even for samples that do not directly use functions in their written program, we still observe a slight improvement. It suggests that our agents are capable of learning from retrieved functions as a reference, and then imitate these functions to write their own programs. For instance, example in Figure~\ref{fig:ex-non-direct} shows the agents learn how to use \texttt{scipy.integrate} by observing the retrieved function \texttt{average\_value\_of\_function(...)}.

%% file: 7_conclusion.tex
\section{Conclusion}
This work proposes \textit{tool-augmented} scientific reasoning, a task aiming to solve challenging scientific problems aided by generalized and scalable tools. To facilitate and evaluate the scientific tool-use abilities of LLMs, we construct a math-related, tool-augmented training corpus \corpus{} and a benchmark \bench{} covering 5 scientific domains. Additionally, we develop open-source agents, \model{} series, as competitive baselines. Extensive experiments reveal that our agents exhibit tool-use abilities exceeding ChatGPT in scientific reasoning tasks.

%% file: limitation.tex
\section*{Limitations}
The primary limitation of our work comes from the way we compile the toolsets in \texttt{SciToolBench}. These tools are constructed directly based on the benchmark's questions, raising concerns about potential information leakage. To address this, we invest significant human effort in our annotation process as detailed in Appendix~\ref{appendix:positive_function}. We manually review and, if necessary, revise all derived functions to ensure their generalizability and quality. As shown in Figure~\ref{fig: retriever} (bottom), our agents achieve only about 40\% accuracy when we provide each question the exact function from which it derives (\ie 100\% hit ratio). It not only highlights the inherent challenge of scientific reasoning tasks, but also suggests that our benchmark suffers minimal impact from the potential information leakage.

We partly attribute this limitation to the absence of a training corpus among scientific (excluding math) domains. The scarcity of annotated solutions for scientific reasoning problems makes it unfeasible to set aside a portion of questions in our benchmark for tool creation. In future work, we plan to collect diverse and high-quality scientific annotations which enable us to develop a more practical and robust tool-augmented benchmark.

\section*{Ethics Statement}
We ensure that \bench{} was constructed in compliance with the terms of use of all source materials and with full respect for the intellectual property and privacy rights of the original authors of the texts. We also provide details on the characteristics and annotation steps of \bench{} in Section~\ref{sec: benchmark} and Appendix~\ref{appendix:details_of_scitoolbench}. We believe our created datasets do not cause any potential risks.

%% file: A_related-work.tex
\section{Detailed Related Work}
\label{appendix:related_work}
\subsection{Scientific Reasoning}
Scientific reasoning can be roughly categorized into two branches: (1) mathematical reasoning and (2) reasoning across other scientific domains.

\paragraph{Mathematical Reasoning.} 
Mathematical (math) reasoning has attracted much more attentions recently. Thanks to abundant training datasets and corpus, there are intensive studies for more powerful math-oriented LLMs by prompt engineering~\cite{qian-etal-2023-creator, zhang2023cumulative, zhou2023solving}, instruction-tuning~\cite{yuan2023scaling, yue2023mammoth, gou2023tora, yu2023metamath, wang2023mathcoder} and even pre-training~\cite{luo2023wizardmath, azerbayev2023llemma, abel}. Regarding instruction-tuning, we notice that recent studies have automatically constructed high-quality instructions from GPT-4, \ie fine-tuning open-source LLMs by Program-of-thought (PoT;~\citealt{chen2022program}) prompting. It enables open-source LLMs to present remarkable performance, even comparable with GPT-4.

\paragraph{Reasoning across Other Domains.} 
There have been intensive works on scientific LLMs~\cite{bran2023chemcrow, jin2023genegpt, fang2023molinstructions} and benchmarks~\cite{hendrycks2021measuring, huang2023ceval, zhang2023m3exam, yue2023mmmu, sun2023scieval}. However, they primarily target on problems involving less complicated reasoning like knowledge retrieval or simple tool utilization. 

Regarding complicated scientific reasoning problems~\cite{chen-etal-2023-theoremqa, wang2023scibench}, questions are scattered among diverse topics and each topic additionally requires domain-specific knowledge. So annotating questions and their solutions domain by domain is much more labor-consuming. Most current benchmarks~\cite{chen-etal-2023-theoremqa, wang2023scibench, zhao2023knowledgemath} merely include hundreds of questions (in all; less for each single domain) from textbooks and provide no training samples. A concurrent work~\cite{zhang2024sciglm} develop a large-scale scientific training corpus, but only focuses three common domains: math, physical and chemistry. Accordingly, the progress of reasoning tasks in these domains is slower than that in math domain: the most competitive approach only achieves $50\%$ and $35\%$ on TheoremQA and SciBench, respectively, not to mention methods built on open-source LLMs. Instead of developing an omniscient and proficient LLMs on reasoning tasks across various scientific domains, we believe it is more practical to teach LLMs the ability to use domain-specific tools to facilitate their reasoning abilities in some domain when external functions (toolset) are attached.

\subsection{Tool Learning}
LLMs, both proprietary ones and open-source ones, demonstrate promising capabilities leveraging external tools to solve problems beyond their limits~\cite{qin2023tool}. Combined with specific tools, these \textit{tool-augmented} LLMs achieve great success on various tasks such as machine learning~\cite{wu2023visual, shen2023hugginggpt, patil2023gorilla, yang2023gpt4tools, liu2023mlbench}, question answering~\cite{peng2023check, gou2023critic}, daily assistance~\cite{xu2023tool, qin2023toolllm, song2023restgpt, gao2023confucius}, \etc.

Previous work usually pre-defines several tools, \eg equation solver or calculator, to facilitate math reasoning tasks~\cite{gou2023critic, lu2023chameleon, hao2023toolkengpt, chen-etal-2023-chatcot, wang2023mint, xu2023lemur, yin2023lumos}. \citet{cai2023large} generalize the concept of tools to \textit{Program functions}. Following this concept, CREATOR~\cite{qian-etal-2023-creator} scale up the function number towards thousand level. However, these ad-hoc, argument-free functions are more like solution wrapper rather than well-generalized tools. CRAFT~\cite{yuan2023craft} targetedly design an automatic pipeline to extract generalized functions for tool-use. 
Though leading to improvement, these functions are still not generalized enough and serve more as reference rather than as tools for direct calling. \citealt{ouyang2023structured} ask LLM to generate chemistry formulae as knowledge reference to assist the following reasoning and achieve enhanced performance on chemistry questions in SciBench. Similar as our attached toolset, \citet{zhao2023knowledgemath} maintain a \textit{knowledge bank} in which saves more than 900 financial definitions/equations/models as the format of functions for retrieval and use. To our best knowledge, our work is the first which (1) finetunes open-source, tool-augmented LLM agents for scientific reasoning tasks and (2) provides a benchmark covering multiple scientific domains to evaluate LLMs' tool-use abilities.

%% file: B_training-details.tex
\section{Training Details}

\subsection{Retriever}
\label{appendix:retriever_training}
To fine-tune a retriever, we construct the training samples from \corpus{}. We concatenate the question and its planning as the query, and view the generated functions as the keys. We finally collect a total of 8603 query-key pairs for training, and split 10\% training samples as validation set.

\begin{align*}
    \text{query} &= [q; G_q] \\
    \text{key} &= f \in \tilde{F}_q
\end{align*}

We follow DPR~\cite{karpukhin-etal-2020-dense} to train a dense retriever $R$. We use \textsc{Roberta-base}~\cite{liu2019roberta} as the backbone. We set the training step as 500, the batch size as 128 and the learning rate as 2e-5. We also set the temperature coefficient of the InfoNCE loss~\cite{oord2019representation} as 0.07. We run this experiment on a single NVIDIA Quadro RTX8000 GPU. The whole training process lasts for about 20 minutes.

\subsection{Planning and Action}
\label{appendix:planning_and_action}
We fine-tune CodeLlamA~\cite{rozière2023code}, Mistral~\cite{jiang2023mistral} and DeepMath~\cite{shao2024deepseekmath} on \corpus{} to develop the planning and action modules in our tool-augmented agents \model{} series, yielding \model{}\textsc{-Coder}, \model{}\textsc{-Mistral} and \model{}\textsc{-DeepMath} respectively. We set the global batch size as 128. We use the learning rate as 2e-5 for CodeLlama, 2e-6 for Mistral, and 5e-6 for DeepMath. We use a cosine scheduler with a 3\% warm-up period for 2 epochs. We train all models with \textit{ZeRO Stage3}~\cite{10.1145/3458817.3476205} on 8 V100 GPUs. The whole training process lasts for about 3 hours for 7B LLMs and 7.5 hours for 13B LLMs.

The planning and action modules share the same model but act differently with different input instructions. We detail the format of planning and action instructions as below:

\noindent \textbf{Planning}. Given a question $q$, we construct a planning sample as $(I_\text{plan}(q), G_q)$, where $I_\text{plan}(q)$ is the input instruction, $G_q$ is the output, and $I_\text{plan}(.)$ is the template for planning module. We provide an example of planning instruction as below:

\begin{lstlisting}[caption={An example of the planning sample. We separate the input instruction and output answer by the dashed line.}, style=prompt]
Read the following question and provide a high-level, step-by-step plan for this problem.
Question: Two complementary angles are in a ratio of $3:2$. What is the measure, in degrees, of the smaller angle?

Answer:
----------------------------------------
To solve this problem, we need to perform the following steps: 
- Set up an equation using the given ratio and the fact that the angles are complementary.
- Solve the equation for the variable representing the ratio.
- Calculate the measure of the smaller angle using the found variable value.
\end{lstlisting}

\noindent \textbf{Action}. Given a question $q$, we construct an action sample as $(I_\text{action}(q, F_q), S_q)$, where $I_\text{action}(q, F_q)$ is the input instruction, $S_q$ is the output answer, and $I_\text{action}(.)$ is the template for action module. $F_q$ are retrieved functions from the toolset in \corpus{}. We adopt the retriever in Appendix~\ref{appendix:retriever_training} to retrieve three functions in the toolset. We provide an example of action instruction as below:

\begin{lstlisting}[caption={An example of the action sample. We separate the input instruction and output answer by the dashed line. We only show one retrieved function in this sample for visualization convenience.}, style=prompt]
Read the following mathematical question and answer it. Please note that you could (optionally, not required) call the following functions when writing the code:

def angle_ratio_to_degrees(ratio):
    """
    Computes the angle measures in degrees given their ratio.

    Parameters:
    - ratio (tuple): The ratio of the angles.

    Returns:
    - tuple: The angle measures in degrees.

    Formula:
    - angle_1 = ratio[0] * x
    - angle_2 = ratio[1] * x
    - angle_3 = ratio[2] * x
    - where x is the common variable such that angle_1 + angle_2 + angle_3 = 180 degrees
    """
    sum_ratios = sum(ratio)
    x = 180 / sum_ratios
    return tuple(angle * x for angle in ratio)

[...Two more retrieved functions...]

Question: Two complementary angles are in a ratio of $3:2$. What is the measure, in degrees, of the smaller angle?

Answer:
----------------------------------------
# Define the ratio of the angles.
ratio = (3, 2)
# Since the angles are complementary, they add up to 90 degrees.
# We can find the value of each part of the ratio by dividing 90 by the sum of the parts of the ratio.
value_of_each_part = 90 / sum(ratio)
# The measure of the smaller angle is the value of each part times the smaller part of the ratio.
smaller_angle = value_of_each_part * min(ratio)
# Print the result.
print(smaller_angle)
\end{lstlisting}

%% file: C_eval-details.tex
\section{Evaluation Details}
\label{appendix:eval_details}
We evaluate (1) the tool-use abilities of LLMs in scientific reasoning task, and (2) their performance when supplemented without any toolsets for comparison. We detail these two settings as below:

\noindent \textbf{With Toolsets.} We use the unified PoT-based prompt~\cite{chen2022program} for all pretraining-based models and our \model{} series. The unified prompt consists of a short task description and two demonstrations. We show the prompt in Appendix~\ref{appendix:evaluation_with_toolsets}. For each question, we provide three retrieved functions and instruct LLMs to use them if (and only if) necessary. Note that we use the same retriever, \ie fine-tuned from \corpus{}, for all LLMs. For MAmmoTH-Coder and ToRA-Coder which are fine-tuned on specific (tool-agnostic) instructions, we try to enable them to use retrieved tools while keeping the formats of their original instructions as much as possible. Specifically, we append a short \textit{tool-augmented} description at the end of their original prompts:

\begin{lstlisting}[style=prompt]
[original prompt]

Please note that you could (optionally, not required) call the following functions when writing the program:

[retrieved functions]
\end{lstlisting}

\noindent \textbf{Without Toolsets.} Similar as above, we use the unified PoT-based prompt~\cite{chen2022program} shown in Appendix~\ref{appendix:evaluation_without_toolsets} for all pretraining-based models and our \model{} series. And we follow the original instructions used for MAmmoTH-Coder and ToRA-Coder to evaluate their performance.

%% file: D_annotation-details.tex
\section{Details of \bench{} Annotation}
\label{appendix:details_of_scitoolbench}

We provide a more thorough description about \bench{} construction in this section. This semi-automatic annotation pipeline involves both GPT-4 and humans to balance the quality and cost. Specifically, we enlist two authors to serve as human annotators. Both of them are graduate students with proficiency in English. Additionally, they hold Bachelor of Science and/or Engineering degrees and have completed undergraduate-level courses in the five scientific domains corresponding to our benchmark. We detail the three subsequent sub-modules in our annotation pipeline, \ie question filtering, positive function construction and negative function construction, as below.

\subsection{Question Filtering}
We curate the questions from TheoremQA~\cite{chen-etal-2023-theoremqa} and SciBench~\cite{wang2023scibench}, both of which are available under the MIT License. Among 1495 questions in these original two datasets, we remove three kinds of questions.

\noindent \textbf{Image-required}: There are 37 questions from TheoremQA which include images and necessitate visual understanding abilities. We remove these samples because our benchmark is text-oriented.

\noindent \textbf{Reasoning-agnostic}: There are some multi-choice questions from TheoremQA which merely requires the memorization of knowledge points but involves little reasoning process. For example:

\begin{tcolorbox}[colback=white, colframe=black, boxrule=1pt, arc=0mm, boxsep=5pt, left=5pt, right=5pt, top=5pt, bottom=5pt]

\noindent \textbf{Question:} The open mapping theorem can be proved by

(a) Baire category theorem.

(b) Cauchy integral theorem.

(c) Random graph theorem.

(d) None of the above. 
\end{tcolorbox}

We manually check each samples and remove 68 such kind of samples.

\noindent \textbf{Over-difficult}: Too hard questions confuse all models and weaken the discrimination of our benchmark. To balance the difficulty and discrimination, we employ 4 advanced proprietary models~\footnote{\texttt{gpt-4}, \texttt{gpt4-32k}, \texttt{gpt-3.5-turbo}, \texttt{gpt-3.5-turbo-16k}} to generate related functions and function-augmented program solutions. We generate 6 solutions for each model (one generated by greedy decoding and the other five by nucleus sampling with 0.6 temperature) and 24 solutions in all. We view questions that are answered incorrectly by all 24 solutions as \textit{over-difficult} questions. We remove all \textit{over-difficult} questions, and retain 73.5\% questions in TheoremQA and 47.8\% in SciBench.

By removing three kinds of samples mentioned above, there are a total of 865 questions in our \bench{} benchmark.

\subsection{Positive Function Construction}
\label{appendix:positive_function}
\noindent \textbf{Function Generation} 

In practice, we merge this sub-module to the process of over-difficult question identification. We randomly sample one set of functions which yield correct solutions for each question. As a result, we collect a total of 1216 candidates for the next verification sub-module. We additionally save other functions leading to correct solutions and use them as reference in the refinement sub-module.

\noindent \textbf{Function Verification}

We verify the generated functions from both correctness and generalizations. We detail them separately as below.

\noindent \textit{1. Correctness:} Since all candidate functions lead to correct solutions, we speculate that almost all of them are correct. We randomly sample 100 functions (20 per domain) and manually check their correctness. The results shown in Table~\ref{tbl:correctness_check} validate our speculation. Therefore, we assume all candidate functions are correct and retain them.

\begin{table}[htbp!]
\small
\centering
\caption{The correctness of 100 randomly sampled functions across five domains.}
    \begin{tabular}{l|cccc}
    \toprule
    & Correct & Partially Correct & Wrong & All \\
    \midrule
    \textbf{Math} & 18 & 2 & 0 & 20\\
    \textbf{Physics} & 19 & 1 & 0 & 20 \\
    \textbf{Chemistry} & 20 & 0 & 0 & 20 \\
    \textbf{Finance} & 19 & 0 & 1 & 20 \\
    \textbf{EECS} &  17 & 3 & 0 & 20 \\
    \midrule
    \textbf{All} & 93 & 6 & 1 & 100 \\
    \bottomrule 
    \end{tabular}
    \label{tbl:correctness_check}
\end{table}

\begin{figure*}[!htbp]
\centering
    \includegraphics[width=0.8\textwidth]{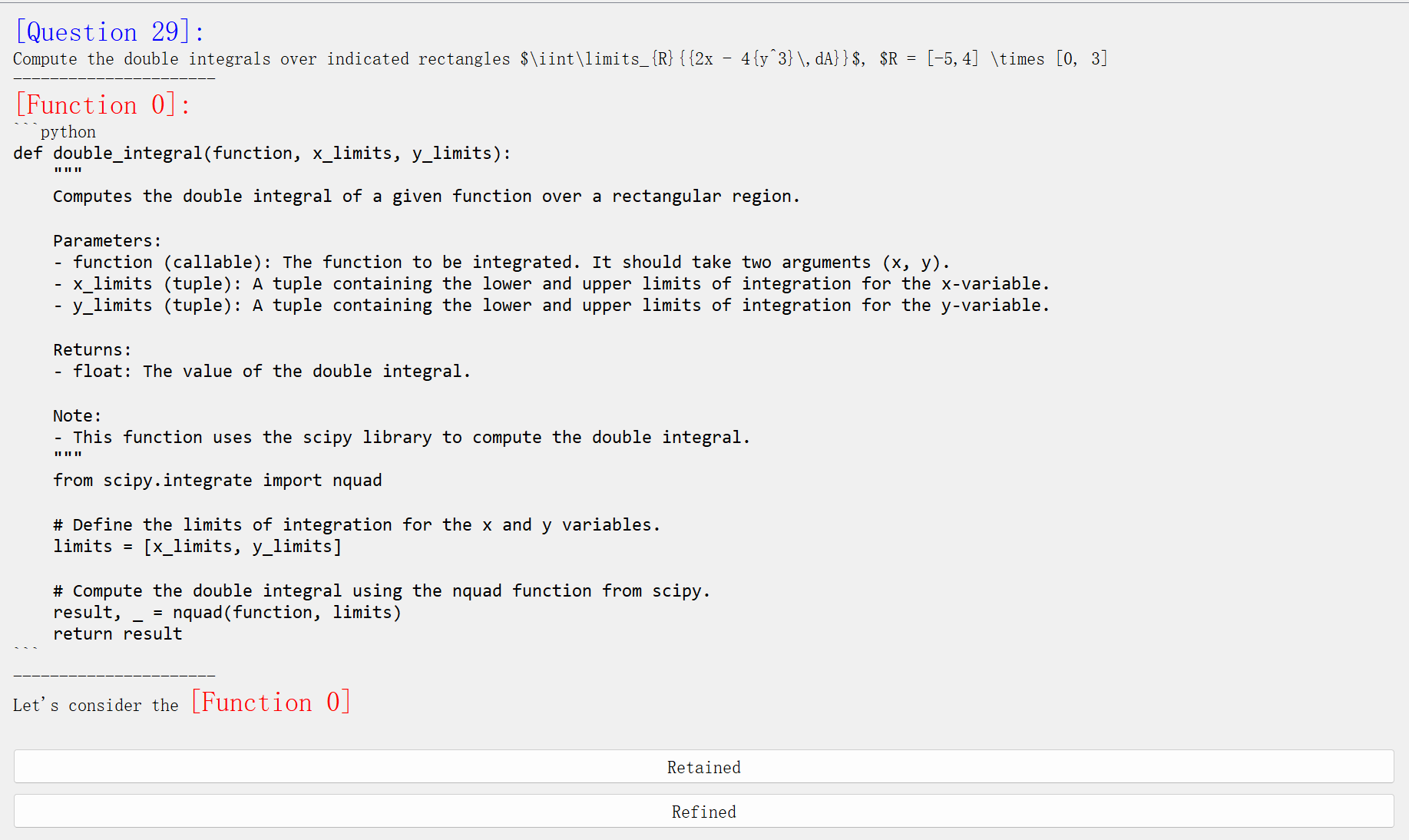}
    \caption{The screenshot of our annotation interface to evaluate functions' generalization.}
    \label{figs:annt_screenshot}
\end{figure*}

\begin{figure*}[!htbp]
\centering
    \includegraphics[width=0.95\textwidth]{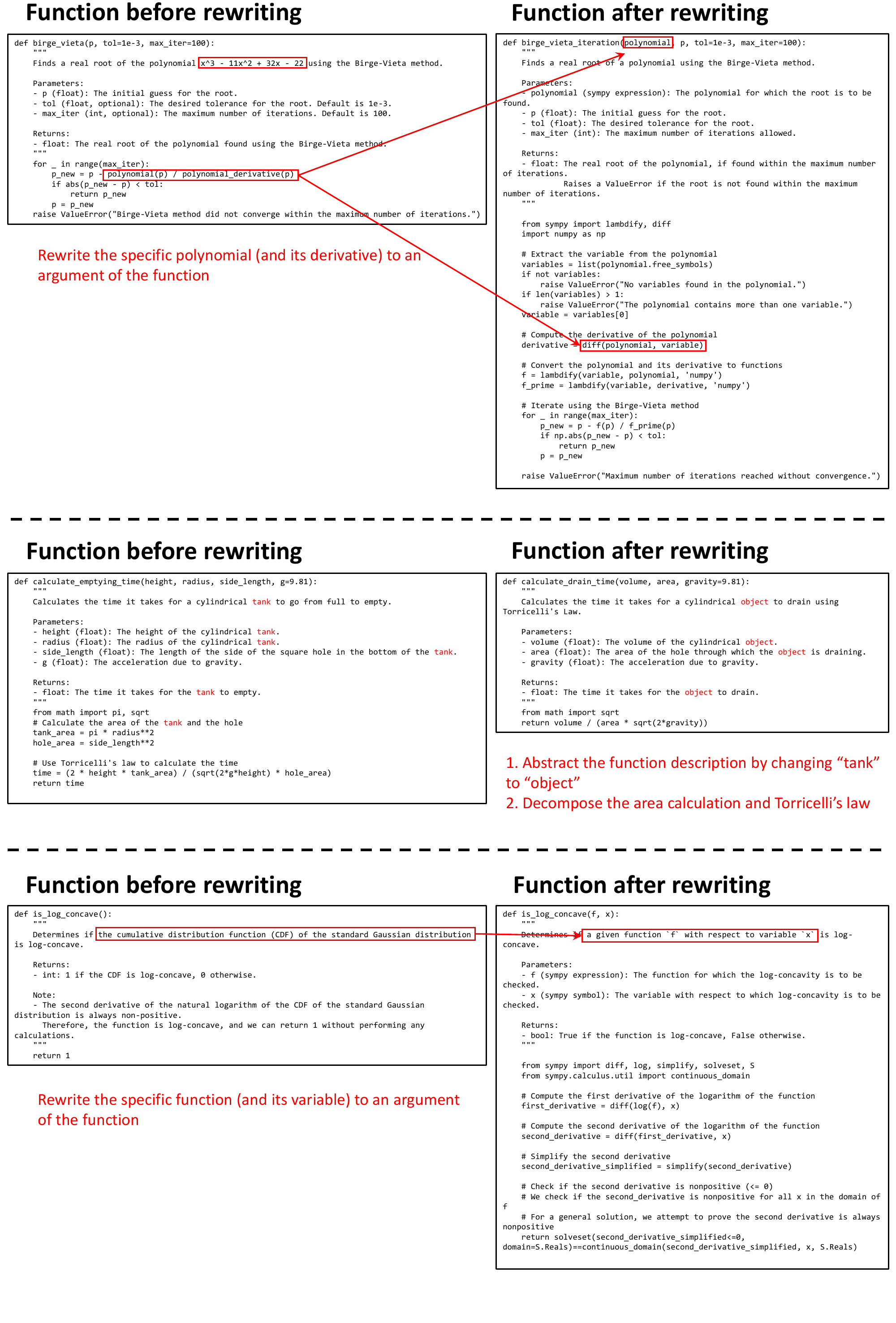}
    \caption{Three examples of \texttt{Refined} functions (before rewriting) and their rewritten functions (after rewriting). We also briefly describe the modifications for each example (color in red).}
    \label{figs:rewrite_function}
\end{figure*}

\noindent \textit{2. Generalization:} We encounter the similar problem as the function construction in \corpus{}, \ie some of the auto-generated functions are not generalized enough. If ad-hoc functions were in the provided toolsets of our benchmark, they would cause a significant overestimation of LLMs' tool-use abilities. To mitigate it as much as possible, we manually check all candidate functions to ensure their generalization. Specifically, we design a binary classification task and assign each function a label in \{\texttt{Retained}, \texttt{Refined}\}. We label a function as \texttt{refined} if it had one of the problems listed below: (1) a pure solution wrapper. (2) merely defining a non-generalized expression (likely only occur in this question). (3) the argument names or document describing the special scenario of corresponding question and not being generalized/abstractive enough. (4) including ad-hoc constants or code snippets. 
The annotators firstly co-annotate 100 functions. We calculate Cohen's kappa value of their annotation results as 0.85, illustrating an ideal agreement. Therefore, the annotators separately annotate the remaining functions. It takes about 6 hours per annotator to classify about 650 functions. We show some \texttt{Refined} function cases in Figure~\ref{figs:rewrite_function}, and the annotation interface in Figure~\ref{figs:annt_screenshot}.

As a result, we collect 1012 \texttt{Retained} and 206 \texttt{Refined} functions. We keep all \texttt{Retained} as the component of positive functions. We also feed the \texttt{Refined} functions to next refinement sub-module to modify them as much as possible. 

\noindent \textbf{Function Refinement} 

\begin{figure*}[!htbp]
\centering
    \includegraphics[width=0.85\textwidth]{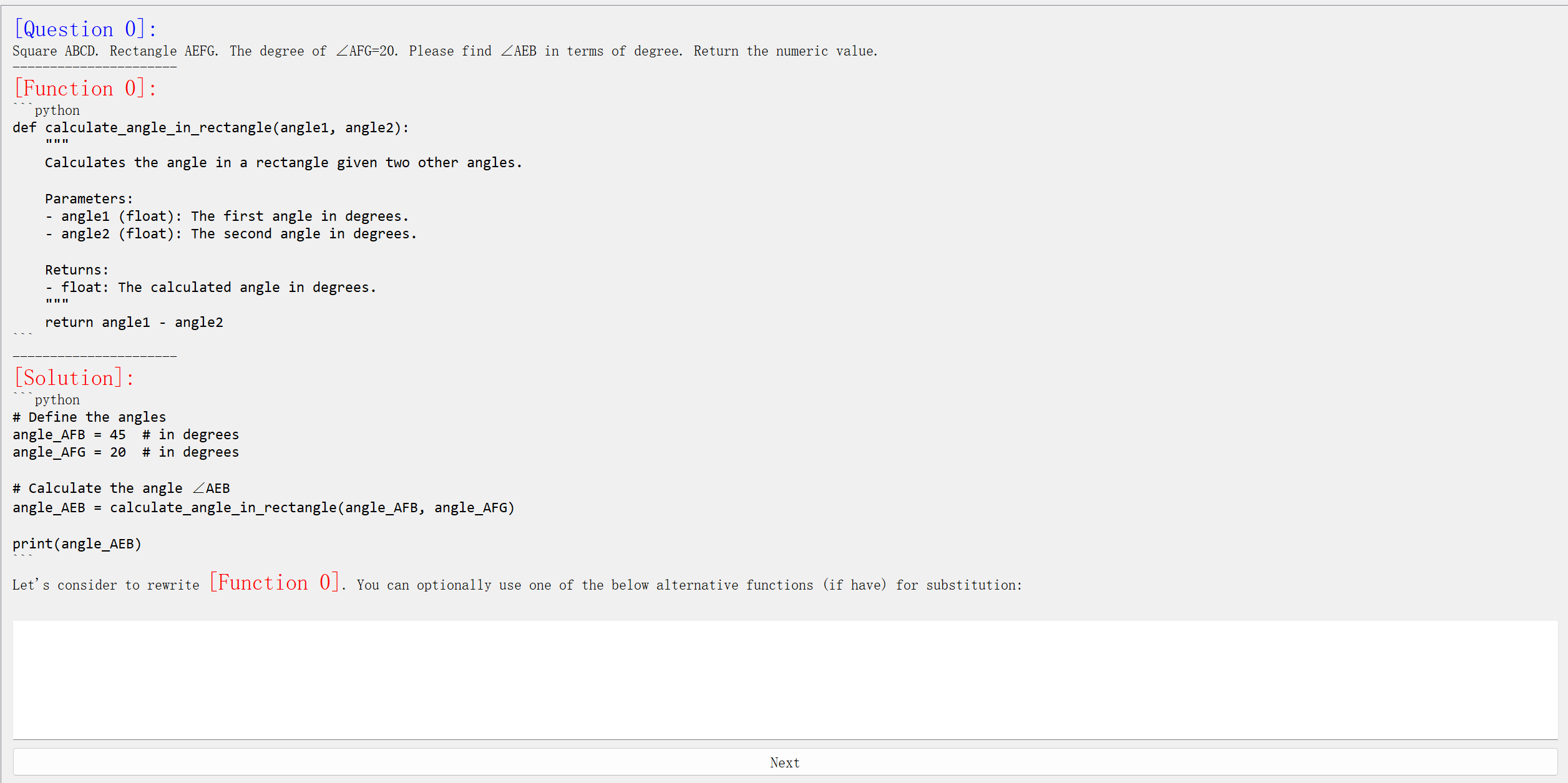}
    \caption{The screenshot of our annotation interface to rewrite functions. We provide no alternative functions in this example for convenience of visualization.}
    \label{figs:rewrite_screenshot}
\end{figure*}

This sub-module aims to rewrite 206 \texttt{Refined} functions to make them qualified. To this end, we associate each function with (1) the question from which it is derived, (2) the function-augmented solutions, and (3) the alternative functions from the generation sub-module (if have). Then we provide them to the annotators. The annotators are asked to rewrite the functions to \textbf{improve their generalization} as much as possible. If one function were successfully rewritten, we also require the annotator to write a solution involving the new function to the related question. The solution must yield correct answer to ensure the correctness of the rewritten function. We show some rewritten cases in Figure~\ref{figs:rewrite_function}, and the screenshot of the annotation interface in Figure~\ref{figs:rewrite_screenshot}.

It takes approximately 12 hours per annotator to check each \texttt{Refined} function and, if applicable, rewrite it. As a consequence, we successfully rewrite 91 \texttt{Refined} functions and drop the remaining ones. We combine these 91 rewritten functions and the 1012 \texttt{Retained} functions to construct 1103 positive functions.

\subsection{Negative Function Construction}
The positive functions constructed above have satisfied the minimum requirements of the toolset in our benchmark. However, we find that such kind of benchmark contains shortcuts for LLM to retrieve and use functions. Take a physical question about \textit{frequency-angular conversion} as example, the previous modules construct a positive function named \texttt{angular\_from\_frequency}(...) to solve this question. Without any other similar functions, the LLMs could readily select and use the \textbf{only} function by superficial shortcuts. These shortcuts significantly weaken the function-understanding and -use abilities evaluation of our benchmark. To mitigate this problem, we design an additional module to eliminate the shortcuts by constructing some (hard) negative functions for each positive function, like \texttt{frequency\_from\_angular}(...) and \texttt{frequency\_from\_energy}(...) in the above example. Among three similar functions, LLMs are forced to understand their usages and choose proper ones to use. In summary, we add negative functions into the toolset to simulate a more challenging scenario and better evaluate LLMs' tool-use abilities.

\label{appendix:neg_func_construct}
\begin{lstlisting}[caption={Prompt for constructing negative functions}, style=prompt]
Given a function about the {subfield} field, could you please write two more functions which satisfy: 
- The functions should be in the same field with the provided function, while the knowledge point is not compulsorily the same. 
- The functions should be similar, but not identical with the provided function. 
- The new written functions should be wrapped as the below format:

New function 1:
```python
[new_written_function_1]
```

New function 2:
```python
[new_written_function_2]
```
\end{lstlisting}

Specifically, we employ GPT-4 for each positive function to generate two similar but not identical functions as the negative functions. The prompt used is shown as below. We do not validate the correctness of negative functions for simplicity, as they are not intended to be used for any question. We filter the duplicated functions and retain the other 1343 functions in all. By merging the 1103 positive functions and 1343 negative functions, we finally collect a total of 2446 functions in our toolset.

\newpage

%% file: E_appendix-case.tex
\section{Examples}
\input{examples/direct_tool_use}
\input{examples/non_direct_tool_use}

%% file: examples/direct_tool_use.tex
\begin{figure}[h!]
    \begin{tcolorbox}[title=Problem \& Planing, left=2mm,right=1mm,top=0mm, bottom=0mm,colback=white]
    \begin{lstlisting}[style=plain]
Read the following mathematical question and answer it. Please note that you could (optionally, not required) call the following functions when writing the code:

Question:Suppose a stock has the following information. It is listed on the London stock exchange and operates throughout Europe. The yield on a UK 10 year treasury is 2.8%. The stock in question will earn 8.6% as per historical data. The Beta for the stock is 1.4, i.e., it is 140% volatile to the changes in the general stock market. What is the expected rate of return?
    \end{lstlisting}
    \tcblower
    \begin{lstlisting}[style=plain]
    (1) Calculate the expected return based on the beta and the yield on a UK 10 year treasury. 
    (2) Add the stock's historical return to the expected return.
    \end{lstlisting}
    \end{tcolorbox}
    \begin{tcolorbox}[title=Retrieved Functions, left=2mm,right=1mm,top=0mm,bottom=0mm, colback=white,colframe=color_tool]
    \begin{lstlisting}[style=python]
def expected_return(rf, beta, rm):
    """
    Computes the expected return using the Capital Asset Pricing Model (CAPM) formula.

    Parameters:
    - rf (float): The risk-free rate.
    - beta (float): The beta of the portfolio.
    - rm (float): The return on the market.

    Returns:
    - float: The expected return.
    """
    return rf + beta * (rm - rf)

def expected_stock_return(true_prob, u, d):
    """
    Calculates the expected return of the stock in a one-period binomial model.
    
    Parameters:
    - true_prob (float): The true probability of the stock price going up.
    - u (float): One plus the rate of capital gain on the stock if the price goes up.
    - d (float): One plus the rate of capital loss on the stock if the price goes down.
    
    Returns:
    - float: The expected return of the stock.
    """
    return true_prob * u + (1 - true_prob) * d

...
    \end{lstlisting}
    \end{tcolorbox}
    \begin{tcolorbox}[title=Solution, left=2mm,right=1mm,top=0mm,bottom=0mm, colback=white,colframe=color_solution]
    \begin{lstlisting}[style=python]
# Given values.
rf = 0.028  # The yield on a UK 10 year treasury
beta = 1.4  # The stock is 140% volatile to the changes in the general stock market
rm = 0.086  # The stock in question will earn 8.6% as per historical data
# Calculate the expected return .
expected_return = expected_return(rf, beta, rm)
# Print the result.
print(expected_return)
    \end{lstlisting}
    \end{tcolorbox}
    \caption{An example of direct function call.}
    \label{fig:ex-direct}
\end{figure}

%% file: examples/non_direct_tool_use.tex
\begin{figure}[htbp!]
    \begin{tcolorbox}[title=Problem \& Planing, left=2mm,right=1mm,top=0.1mm, bottom=0.1mm,colback=white]
    \begin{lstlisting}[style=plain]
Read the following mathematical question and answer it. Please note that you could (optionally, not required) call the following functions when writing the code:

Question:The linear density in a rod $8 \mathrm{~m}$ long is $12 / \sqrt{x+1} \mathrm{~kg} / \mathrm{m}$, where $x$ is measured in meters from one end of the rod. Find the average density of the rod.
    \end{lstlisting}
    \tcblower
    \begin{lstlisting}[style=plain]
    (1) Integrate the linear density function with respect to x from 0 to 8. 
    (2) Divide the result by the length of the rod.
    \end{lstlisting}
    \end{tcolorbox}
    \begin{tcolorbox}[title=Retrieved Functions, left=2mm,right=1mm,top=0.1mm,bottom=0mm, colback=white,colframe=color_tool]
    \begin{lstlisting}[style=python]
def average_value_of_function(f, a, b):
    """
    Calculates the average value of a function `f` on the interval `[a, b]`.
    Parameters: - f (sympy expression): The function for which the average value is to be found.
    - a (float): The lower limit of the interval.
    - b (float): The upper limit of the interval.
    Returns: float: The average value of the function on the interval.
    """
    from sympy import integrate, symbols
    x = symbols('x')
    integral = integrate(f, (x, a, b))
    average_value = integral / (b - a)
    return average_value

def avg_distance(vectors):
    """
    Finds the average distance from the origin that can be reached by adding and subtracting the given vectors.
    Parameters: - vectors (list of numpy arrays): The vectors.
    Returns: float: The average distance.
    """
    from itertools import product
    import numpy as np
    total_distance = 0
    count = 0
    for signs in product([-1, 1], repeat=len(vectors)):
        resultant = sum(v * s for v, s in zip(vectors, signs))
        distance = np.linalg.norm(resultant)
        total_distance += distance
        count += 1
    return total_distance / count

...
    \end{lstlisting}
    \end{tcolorbox}
    \begin{tcolorbox}[title=Solution, left=2mm,right=1mm,top=0.1mm,bottom=0mm, colback=white,colframe=color_solution]
    \begin{lstlisting}[style=python]
# The linear density is given by 12 / sqrt(x+1) kg/m.
# The average density is the total mass of the rod divided by its length.
# The total mass of the rod is the integral of the linear density times the length of the rod from 0 to 8.
from sympy import symbols, integrate
x = symbols('x')
length = 8
# Define the linear density function.
linear_density = 12 / (x + 1)**0.5
# Calculate the total mass of the rod.
total_mass = integrate(linear_density, (x, 0, length))
# Calculate the average density by dividing the total mass by the length.
average_density = total_mass / length
print(average_density)
    \end{lstlisting}
    \end{tcolorbox}
    \caption{An example of implicitly using functions as a reference.}
    \label{fig:ex-non-direct}
\end{figure}

%% file: F_prompt.tex
\clearpage
\section{Prompts}
\subsection{Tool/Function Collection}
\label{appendix:function_collection}
\begin{lstlisting}[caption={Prompt used for planning and function collection}, style=prompt]
Given the question and the reference solution, do the following things: 

- Think about what math knowledge points are required to solve this problem step by step.  
- write some python one or more functions to abstract the solution. Please note that the functions should be well-documented as much as possible and not too specific (for example, do not write the values in this problem within the functions. Pass them as the function arguments). We hope your written functions could be re-used in anywhere else.
-Instantiate these functions to solve the problem. The last line of your program should be a 'print' command to print the final answer

Here are some examples you may refer to:

Question: There are integers $b,c$ for which both roots of the polynomial $x^2-x-1$ are also roots of the polynomial $x^5-bx-c$. Determine the product $bc$.
Answer: Let $r$ be a root of $x^2-x-1$. Then, rearranging, we have\n$$r^2 = r+1.$$Multiplying both sides by $r$ and substituting gives\n\\begin{align*}\nr^3 &= r^2+r \\\\\n&= (r+1)+r \\\\\n&= 2r+1.\n\\end{align*}Repeating this process twice more, we have\n\\begin{align*}\nr^4 &= r(2r+1) \\\\\n&= 2r^2+r \\\\\n&= 2(r+1)+r \\\\\n&= 3r+2\n\\end{align*}and\n\\begin{align*}\nr^5 &= r(3r+2) \\\\\n&= 3r^2+2r \\\\\n&= 3(r+1)+2r \\\\\n&= 5r+3.\n\\end{align*}Thus, each root of $x^2-x-1$ is also a root of $x^5-5x-3$, which gives $bc = 5\\cdot 3 = \\boxed{15}$.
Think: To solve this question, we can follow the steps below: (1) Find the roots of the polynomial $x^2-x-1$. (2) Substitute them into the the polynomial $x^5-bx-c$ and obtain two equations. (3) Solve the equations.
Functions:
```function 1
def find_roots_of_polynomial(polynomial, variable):
    """
    Finds the roots of a given polynomial using the sympy library.

    Parameters:
    - polynomial (sympy expression): The polynomial whose roots are to be found.
    - variable (sympy symbol): The variable of the polynomial.

    Returns:
    - list: The roots of the polynomial.
    """

    from sympy import solve
    roots = solve(polynomial, variable)
    return roots
```

```function 2
def substitute_roots_into_polynomial(roots, polynomial, variable):
    """
    Substitutes the given roots into the polynomial and returns the resulting expressions.

    Parameters:
    - roots (list): The list of roots to be substituted into the polynomial.
    - polynomial (sympy expression): The polynomial into which the roots are to be substituted.
    - variable (sympy symbol): The variable of the polynomial.

    Returns:
    - list: The resulting expressions after substituting the roots into the polynomial.
    """

    return [polynomial.subs(variable, root) for root in roots]
```

```function 3
def solve_equations(equations, variables):  
    """
    Solves a system of equations for the specified variables using the sympy library.

    Parameters:
    - equations (list of sympy expressions or a single sympy expression): 
        The equations to be solved. If solving a single equation, this can be a single expression.
    - variables (list of sympy symbols or a single sympy symbol): 
        The variables for which the solution is to be found. If solving for a single variable, this can be a single symbol.

    Returns:
    - list of dictionaries: Each dictionary represents a solution, with keys being the variables and values being their corresponding values. 
      If there's only one solution, the list will contain a single dictionary.
    """

    from sympy import solve  
    solution = solve(equations, variables, dict=True)  
    return solution  
```

Solution:
```python
# Import required functions and classes from sympy
from sympy import symbols, Eq

# Define the variable and the polynomials
x, b, c = symbols('x b c')
polynomial1 = x**2 - x - 1
polynomial2 = x**5 - b*x - c

# Find the roots of the first polynomial
roots = find_roots_of_polynomial(polynomial1, x)

# Substitute the roots into the second polynomial
resulting_expressions = substitute_roots_into_polynomial(roots, polynomial2, x)

# Set up the equations based on the resulting expressions
equations = [Eq(expr, 0) for expr in resulting_expressions]

# Solve the system of equations for b and c
solutions = solve_equations(equations, (b, c))
# This linear system has only one solution
solution = solutions[0]

# Calculate the product bc
product_bc = solution[b] * solution[c]
print(product_bc)
```

---

Question: Medians $\\overline{DP}$ and $\\overline{EQ}$ of $\\triangle DEF$ are perpendicular.  If $DP= 18$ and $EQ = 24$, then what is ${DE}$?
Answer: Point $G$ is the centroid of $\\triangle DEF$, so $DG:GP = EG:GQ = 2:1$.  Therefore, $DG = \\frac23(DP) = 12$ and $EG = \\frac23(EQ) =16$, so applying the Pythagorean Theorem to $\\triangle EGD$ gives us $DE = \\sqrt{EG^2 + GD^2} = \\boxed{20}$.
Think: Given two perpendicular medians in a triangle, we need to perform the following steps: (1) Identify the relationship between the segments of medians and the centroid. (2) Use the ratios provided to determine the lengths of the individual segments from the centroid to the vertices. (3) Use the Pythagorean theorem to determine the length of the side connecting the two vertices from which the medians originate.
Functions:
```function 1
def median_segments_length(median_length, ratio):
    """
    Computes the lengths of the segments of a median split by the centroid.

    Parameters:
    - median_length (float): Total length of the median.
    - ratio (tuple): Ratio in which the centroid splits the median. Default is (2,1) for standard triangles.

    Returns:
    - tuple: Lengths of the two segments.

    Formula:
    - segment_1 = ratio[0]/sum(ratio) * median_length
    - segment_2 = ratio[1]/sum(ratio) * median_length
    """
    segment_1 = ratio[0] / sum(ratio) * median_length
    segment_2 = ratio[1] / sum(ratio) * median_length
    return segment_1, segment_2
```

```function 2
def pythagorean_theorem(a, b):
    """
    Computes the hypotenuse of a right triangle given two legs.

    Parameters:
    - a, b (float): Lengths of the two legs.

    Returns:
    - float: Length of the hypotenuse.

    Formula:
    - c = sqrt(a^2 + b^2)
    """
    from sympy import sqrt
    return sqrt(a**2 + b**2)
```

Solution:
```python
# Given values
DP = 18
EQ = 24

# Point $G$ is the centroid.
ratio = (2,1)
# Determine the lengths of the segments split by the centroid
DG, GP = median_segments_length(DP, ratio)
EG, GQ = median_segments_length(EQ, ratio)

# Use Pythagoras theorem to determine DE
DE = pythagorean_theorem(EG, DG)
print(DE)
```

\end{lstlisting}

\subsection{Self-rectification}
\label{appendix:self_rectification}
\begin{lstlisting}[caption={Prompt used for self-rectification}, style=prompt]

{
    "role": "system", 
    "content": "Suppose you are writing Python programs to solve mathematical questions. Your original program, however, can not run normally for some reasons. Given the error message returned from the interpreter, please debug your program carefully and generate the new program. Your generated correction MUST FOLLOW the following format:\n```function 1\n[the first created function]\n```\n...\n```function N\n[the N-th created function]\n```\n```python\n[solution]\n```.\nDO NOT only quote your output as ```python\n[solution]\n```"
},
{
    "role": "user",
    "content": "Original Program:\n```function 1\ndef solve_quadratic(a, b, c):\n    from sympy import symbols, Eq, solve\n    x = symbols('x')\n    quadratic = Eq(a * x**2 + b * x + c, 0)\n    solutions = solve(quadratic, x)\n    return solutions\n```\n\n```python\ndef find_d_values():\n    a, b, c, d = symbols('a b c d')\n    eq1 = Eq(6 * a * d, a + d)\n    eq2 = Eq((6 * d - 1) * (1 - a), a + d)\n    eq3 = Eq(a, 4 * d - 1)\n    solution = solve([eq1, eq2, eq3], (a, b, c, d), dict=True)\n    d_values = [sol[d] for sol in solution]\n    return d_values\nd_values = find_d_values()\nprint(d_values)\n```\n\nError Message:\nFile <string>:15\n\nFile <string>:9, in find_d_values()\n\nNameError: name 'Eq' is not defined"
},
{
    "role": "assistant",
    "content": "Analysis: The error is occurring because the Eq function from the sympy module is not recognized in the find_d_values function. This is happening because the ```from sympy import symbols, Eq, solve``` line is only inside the ```solve_quadratic``` function, so Eq and other imported functions and classes are only available within the scope of that function. To fix this, you should  claim import statement within the ```find_d_values``` function as well.\n\nNew Program:\n```function 1\ndef solve_quadratic(a, b, c):\n    from sympy import symbols, Eq, solve\n    x = symbols('x')\n    quadratic = Eq(a * x**2 + b * x + c, 0)\n    solutions = solve(quadratic, x)\n    return solutions\n```\n\n```python\ndef find_d_values():\n    from sympy import symbols, Eq, solve\n    a, b, c, d = symbols('a b c d')\n    eq1 = Eq(6 * a * d, a + d)\n    eq2 = Eq((6 * d - 1) * (1 - a), a + d)\n    eq3 = Eq(a, 4 * d - 1)\n    solution = solve([eq1, eq2, eq3], (a, b, c, d), dict=True)\n    d_values = [sol[d] for sol in solution]\n    return d_values\nd_values = find_d_values()\nprint(d_values)\n```"
},
{
    "role": "user",
    "content": "finished"
},
{
    "role": "user",
    "content": "Original Program:\n```function 1\ndef matrix_square(matrix):\n    from sympy import Matrix\n    if not isinstance(matrix, Matrix):\n        matrix = Matrix(matrix)\n    return matrix * matrix\n```\n\n```function 2\ndef solve_equations(equations, variables):  \n    from sympy import solve  \n    solution = solve(equations, variables, dict=True)  \n    return solution  \n```\n\n```function 3\ndef count_solutions(solutions):\n    return len(solutions)\n```\n\n```python\nfrom sympy import symbols, Eq  \na, d = symbols('a d')  \nmatrix = [[a, 3], [-8, d]]  \nsquared_matrix = matrix_square(matrix)  \nidentity_matrix = [[1, 0], [0, 1]]  \nequations = [Eq(squared_matrix[i][j], identity_matrix[i][j]) for i in range(2) for j in range(2)]  \nsolution = solve_equations(equations, [a, d])  \nans = count_solutions(solution)\nprint(ans)\n```\n\nError Message:\nFile <string>:17\n\nFile <string>:17, in <listcomp>(.0)\n\nTypeError: 'Add' object is not subscriptable"
},
{
    "role": "assistant",
    "content": "Analysis: The error message is occurring because when you are creating the list of equations with the list comprehension, squared_matrix[i][j] is not working as expected. The squared_matrix is a SymPy Matrix object, and you should use the .row and .col methods.\n\nNew Program:\n```function 1\ndef matrix_square(matrix):\n    from sympy import Matrix\n    if not isinstance(matrix, Matrix):\n        matrix = Matrix(matrix)\n    return matrix * matrix\n```\n\n```function 2\ndef solve_equations(equations, variables):  \n    from sympy import solve  \n    solution = solve(equations, variables, dict=True)  \n    return solution  \n```\n\n```function 3\ndef count_solutions(solutions):\n    return len(solutions)\n```\n\n```python\nfrom sympy import symbols, Eq  \na, d = symbols('a d')  \nmatrix = [[a, 3], [-8, d]]  \nsquared_matrix = matrix_square(matrix)  \nidentity_matrix = [[1, 0], [0, 1]]  \nequations = [Eq(squared_matrix.row(i)[j], identity_matrix[i][j]) for i in range(2) for j in range(2)]   \nsolution = solve_equations(equations, [a, d])  \nans = count_solutions(solution)\nprint(ans)\n```"
},
{
    "role": "user",
    "content": "finished"
},
{
    "role": "user",
    "content": "Original Program:\n```function 1\ndef sum_to_product_cos(alpha, beta):\n    return (cmath.cos(alpha - beta) + cmath.cos(alpha + beta)) / 2\n```\n\n```function 2\ndef simplify_expression(expression):  \n    from sympy import simplify  \n    return simplify(expression)  \n```\n\n```python\nfrom sympy import symbols, cos\n\nx = symbols('x')\nexpr = cos(x) + cos(3*x) + cos(7*x) + cos(9*x)\n\n# Apply sum-to-product formula\nexpr1 = 2 * sum_to_product_cos(5*x, 4*x)\nexpr2 = 2 * sum_to_product_cos(5*x, 2*x)\n\n# Simplify the expression\nsimplified_expr = simplify_expression(expr1 + expr2)\n\n# Extract the values of a, b, c, and d and sum them up\na, b, c, d = 4, 1, 3, 5\nans = a + b + c + d\nprint(ans)\n```\n\nError Message:\nFile <string>:12\n\nFile <string>:2, in sum_to_product_cos(alpha, beta)\n\nNameError: name 'cmath' is not defined"
},
{
    "role": "assistant",
    "content": "Analysis: The error message is occurring because the cmath module is being used in the sum_to_product_cos function but it has not been imported. You can fix this by importing the cmath module in this function. New Program:\n```function 1\ndef sum_to_product_cos(alpha, beta):\n    import cmath\n    return (cmath.cos(alpha - beta) + cmath.cos(alpha + beta)) / 2\n```\n\n```function 2\ndef simplify_expression(expression):  \n    from sympy import simplify  \n    return simplify(expression)\n```  \n\n```python\nfrom sympy import symbols, cos\n\nx = symbols('x')\nexpr = cos(x) + cos(3*x) + cos(7*x) + cos(9*x)\n\n# Apply sum-to-product formula\nexpr1 = 2 * sum_to_product_cos(5*x, 4*x)\nexpr2 = 2 * sum_to_product_cos(5*x, 2*x)\n\n# Simplify the expression\nsimplified_expr = simplify_expression(expr1 + expr2)\n\n# Extract the values of a, b, c, and d and sum them up\na, b, c, d = 4, 1, 3, 5\nans = a + b + c + d\nprint(ans)\n```"
},
{
    "role": "user",
    "content": "Error message:\nTraceback (most recent call last):\n  File, line 9, in <module>\n    exec(\n  File '<string>', line 13, in <module>\n  File '<string>', line 3, in sum_to_product_cos\n  File, line 356, in __complex__\n    return complex(float(re), float(im))\n  File, line 351, in __float__\n    raise TypeError('Cannot convert expression to float')\nTypeError: Cannot convert expression to float\n"
},
{
    "role": "assistant",
    "content": "Analysis: The error occurs because you are trying to use cmath.cos function (which expects float or complex numbers as arguments) with SymPy symbolic expressions (5x and 4x, 5x and 2x are symbolic expressions). To fix this, you should use SymPy's cos function instead of cmath.cos function to work with symbolic expressions.\n\nNew Program:\ndef sum_to_product_cos(alpha, beta):\n    from sympy import cos\n    return (cos(alpha - beta) + cos(alpha + beta)) / 2\ndef simplify_expression(expression):  \n    from sympy import simplify  \n    return simplify(expression)  \nfrom sympy import symbols, cos\n\nx = symbols('x')\nexpr = cos(x) + cos(3*x) + cos(7*x) + cos(9*x)\n\n# Apply sum-to-product formula\nexpr1 = 2 * sum_to_product_cos(5*x, 4*x)\nexpr2 = 2 * sum_to_product_cos(5*x, 2*x)\n\n# Simplify the expression\nsimplified_expr = simplify_expression(expr1 + expr2)\n\n# Extract the values of a, b, c, and d and sum them up\na, b, c, d = 4, 1, 3, 5\nans = a + b + c + d\nprint(ans)"
},
{
    "role": "user",
    "content": "finished"
},
{
    "role": "user",
    "content": "Original Program:\n```function 1\ndef tan_double_angle(tan_theta):\n    return 2 * tan_theta / (1 - tan_theta**2)\n```\n\n```function 2\ndef tan_triple_angle(tan_theta):\n    return (3 * tan_theta - tan_theta**3) / (1 - 3 * tan_theta**2)\n```\n\n```function 3\ndef solve_polynomial(polynomial, x):\n    from sympy import solve\n    return solve(polynomial, x)\n```\n\n```python\nfrom sympy import symbols\ntan_theta = symbols('tan_theta')\ntan_2_theta = tan_double_angle(tan_theta)\ntan_3_theta = tan_triple_angle(tan_theta)\nequation = tan_theta + tan_2_theta + tan_3_theta\nsimplified_equation = simplify_expression(equation)\ntan_theta_solutions = solve_polynomial(simplified_equation, tan_theta)\n\n# Find the solution in the given range (0 < tan_theta < 1)\ntan_theta_value = [solution for solution in tan_theta_solutions if 0 < solution < 1][0]\nprint(tan_theta_value)\n```\n\nError Message:\nFile <string>:13\n\nNameError: name 'simplify_expression' is not defined"
},
{
    "role": "assistant",
    "content": "Analysis: The error message is occurring because there is no function named simplify_expression defined in your script. To fix this, you can define such a function that uses SymPy's simplify function to simplify expressions.\n\nNew Program:\n```function 1\ndef simplify_expression(expression):  \n    from sympy import simplify  \n    return simplify(expression)\n```\n\n```function 2\ndef tan_double_angle(tan_theta):\n    return 2 * tan_theta / (1 - tan_theta**2)\n```\n\n```function 3\ndef tan_triple_angle(tan_theta):\n    return (3 * tan_theta - tan_theta**3) / (1 - 3 * tan_theta**2)\n```\n\n```function 4\ndef solve_polynomial(polynomial, x):\n    from sympy import solve\n    return solve(polynomial, x)\n```\n\n```python\nfrom sympy import symbols\ntan_theta = symbols('tan_theta')\ntan_2_theta = tan_double_angle(tan_theta)\ntan_3_theta = tan_triple_angle(tan_theta)\nequation = tan_theta + tan_2_theta + tan_3_theta\nsimplified_equation = simplify_expression(equation)\ntan_theta_solutions = solve_polynomial(simplified_equation, tan_theta)\n\n# Find the solution in the given range (0 < tan_theta < 1)\ntan_theta_value = [solution for solution in tan_theta_solutions if 0 < solution < 1][0]\nprint(tan_theta_value)\n```"
},
{
    "role": "user",
    "content": "finished"
}
\end{lstlisting}

\subsection{Function-augmented Solutions}
\label{appendix:cross_retrieval}
\begin{lstlisting}[caption={Prompt used for the generation of function-augmented solutions (cross-retrieval strategy)}, style=prompt]
You will encounter a mathematical problem and are required to write a piece of Python code to solve this problem.

Now we have a suite of wrapped functions. Take note:
- The newly provided wrapped functions have NOT been verified. They may be irrelevant or potentially flawed.
- It's essential that the solution doesn't overly depend on wrapped functions. 
  You're welcome to utilize one or more functions from the new set in your solution but only after you've determined: 
  (1) Their accuracy. 
  (2) Their inclusion significantly streamlines the problem-solving approach. 

Additionally take note that 
    (1) The last line of your written code shall be a 'print' command to print the final answer.
    (2) The wrapped functions should not be duplicated within your code. Instead, call them directly if needed.
    (3) Should you need to create custom functions, do so without adding documentation comments for the sake of brevity.
    (4) Write simple but clear annotations interleaving your code solution.

"""
Retrieved functions:
[List of called function names from the new set]

```python
[Your Written Python Code.]
```
"""

For example:
---
Question: What is the 100th digit to the right of the decimal point in the decimal representation of $\frac{13}{90}$?

New provided functions:
```New Function 0
def decimal_representation(numerator, denominator, max_digits=1000):
    """
    Computes the decimal representation of a fraction.

    Parameters:
    - numerator (int): The numerator of the fraction.
    - denominator (int): The denominator of the fraction.
    - max_digits (int): The maximum number of decimal digits to compute.

    Returns:
    - str: The decimal representation of the fraction as a string.
    """

    result = ""
    remainder = numerator % denominator
    for _ in range(max_digits):
        remainder *= 10
        result += str(remainder // denominator)
        remainder %= denominator
        if remainder == 0:
            break
    return result
```

```New Function 1
def decimal_to_scientific(decimal_number):
    from sympy import log, floor
    exponent = -floor(log(decimal_number, 10))
    coefficient = decimal_number * 10**(-exponent)
    return coefficient, exponent
```

```New Function 2
def repeating_decimal_representation(numerator, denominator):
    """
    Computes the repeating decimal representation of a fraction.

    Parameters:
    - numerator (int): The numerator of the fraction.
    - denominator (int): The denominator of the fraction.

    Returns:
    - str: The repeating decimal representation of the fraction as a string.
    """

    # Initialize the result string and a dictionary to store remainders.
    result = ""
    remainders = {}

    # Perform long division to find the decimal representation.
    while numerator != 0:
        # If the remainder has been seen before, we found the repeating block.
        if numerator in remainders:
            start = remainders[numerator]
            return result[:start] + "(" + result[start:] + ")"
        # Otherwise, store the remainder and continue the division.
        remainders[numerator] = len(result)
        numerator *= 10
        result += str(numerator // denominator)
        numerator %= denominator

    return result
```

```New Function 3
def nth_digit_of_decimal_representation(numerator, denominator, n):
    """
    Computes the nth digit after the decimal point of the decimal representation of a fraction.

    Parameters:
    - numerator (int): The numerator of the fraction.
    - denominator (int): The denominator of the fraction.
    - n (int): The position of the digit after the decimal point.

    Returns:
    - int: The nth digit after the decimal point of the decimal representation of the fraction.
    """

    # Get the repeating decimal representation of the fraction.
    decimal_representation = repeating_decimal_representation(numerator, denominator)

    # Remove the parentheses from the repeating block.
    decimal_representation = decimal_representation.replace("(", "").replace(")", "")

    # Calculate the nth digit using the repeating block.
    return int(decimal_representation[(n - 1) % len(decimal_representation)])
```

Retrieved functions:
[decimal_representation, nth_digit_of_decimal_representation]

```python
# Use the nth_digit_of_decimal_representation function to find the 100th digit
numerator = 13
denominator = 90
n = 100

# Call the function and print the result
result = nth_digit_of_decimal_representation(numerator, denominator, n)
print(result)
```

---
Question: The square root of $x$ is greater than 3 and less than 4. How many integer values of $x$ satisfy this condition?

New provided functions:
```New Function 0
def solve_square_root_equation(a, b, c):
    """
    Solves a square root equation of the form sqrt(ax - b) = c.

    Parameters:
    - a (float): Coefficient of x inside the square root.
    - b (float): Constant term inside the square root.
    - c (float): Constant term on the right side of the equation.

    Returns:
    - float: The value of x that satisfies the equation.

    Formula:
    - x = (c^2 + b) / a
    """
    return (c**2 + b) / a
```

```New Function 1
def find_integer_square_less_than_double():
    """
    Finds the only integer whose square is less than its double.

    Returns:
    - int: The integer that satisfies the condition.

    Method:
    - Iterate through integers starting from 1, and check if the square of the integer is less than its double.
    - If the condition is satisfied, return the integer.
    - If the condition is not satisfied for any integer up to a certain limit, return None.
    """
    limit = 100
    for x in range(1, limit):
        if x**2 < 2*x:
            return x
    return None
```

```New Function 2
def solve_equation():
    """
    Solves the equation (x-2)^(25-x^2) = 1 for integer solutions.

    Returns:
    - list: A list of integer solutions for x.
    """
    solutions = []

    # Case 1: Exponent is 0 (25 - x^2 = 0)
    x1 = 5
    x2 = -5
    solutions.extend([x1, x2])

    # Case 2: Base is 1 (x - 2 = 1)
    x3 = 3
    solutions.append(x3)

    # Case 3: Base is -1 and exponent is even (x - 2 = -1 and 25 - x^2 = 2n for some integer n)
    x4 = 1
    solutions.append(x4)

    return solutions
```

```New Function 3
def count_integers_in_range(lower_bound, upper_bound, exclude_zero=True):
    """
    Counts the number of integers within a given range.

    Parameters:
    - lower_bound (int): The lower bound of the range.
    - upper_bound (int): The upper bound of the range.
    - exclude_zero (bool): Whether to exclude 0 from the count. Default is True.

    Returns:
    - int: The number of integers within the range.
    """
    count = upper_bound - lower_bound + 1
    if exclude_zero and lower_bound <= 0 and upper_bound >= 0:
        count -= 1
    return count
```

Retrieved functions:
[]

```python
# The lower and upper bounds of x for which sqrt(x) > 3 and sqrt(x) < 4
lower_bound = 9
upper_bound = 16

# Counting the number of integers between 9 (exclusive) and 16 (exclusive)
num_integers = len([x for x in range(lower_bound + 1, upper_bound)])

# Printing the result
print(num_integers)
```
\end{lstlisting}

\subsection{Evaluation with Toolsets}
\label{appendix:evaluation_with_toolsets}
\begin{lstlisting}[caption={Prompt used for evaluation (setting with toolsets)}, style=prompt]
Read the following questions and answer them. For each question, you are required to write a Python program to solve it.
Please note that we provide you several functions for each question. You could (optionally, not required) call the functions to help you to solve the question if necessary.
Note that the last line of your program should be a 'print' command to print the final answer

----------------------------------------------------
Question:
What is the 100th digit to the right of the decimal point in the decimal representation of $\\frac{13}{90}$?

Functions:
def repeating_decimal_representation(numerator, denominator):
    """
    Computes the repeating decimal representation of a fraction.
                                    
    Parameters:
    - numerator (int): The numerator of the fraction.
    - denominator (int): The denominator of the fraction.

    Returns:
    - str: The repeating decimal representation of the fraction as a string.
    """

    # Initialize the result string and a dictionary to store remainders.
    result = ""
    remainders = {}

    # Perform long division to find the decimal representation.
    while numerator != 0:
        # If the remainder has been seen before, we found the repeating block.
        if numerator in remainders:
            start = remainders[numerator]
            return result[:start] + "(" + result[start:] + ")"
        # Otherwise, store the remainder and continue the division.
        remainders[numerator] = len(result)
        numerator *= 10
        result += str(numerator // denominator)
        numerator %= denominator

    return result

def nth_digit_of_decimal_representation(numerator, denominator, n):
    """
    Computes the nth digit after the decimal point of the decimal representation of a fraction.

    Parameters:
    - numerator (int): The numerator of the fraction.
    - denominator (int): The denominator of the fraction.
    - n (int): The position of the digit after the decimal point.

    Returns:
    - int: The nth digit after the decimal point of the decimal representation of the fraction.
    """

    # Get the repeating decimal representation of the fraction.
    decimal_representation = repeating_decimal_representation(numerator, denominator)

    # Remove the parentheses from the repeating block.
    decimal_representation = decimal_representation.replace("(", "").replace(")", "")

    # Calculate the nth digit using the repeating block.
    return int(decimal_representation[(n - 1) % len(decimal_representation)])

def decimal_representation(numerator, denominator, max_digits=1000):
    """
    Computes the decimal representation of a fraction.

    Parameters:
    - numerator (int): The numerator of the fraction.
    - denominator (int): The denominator of the fraction.
    - max_digits (int): The maximum number of decimal digits to compute.

    Returns:
    - str: The decimal representation of the fraction as a string.
    """

    result = ""
    remainder = numerator % denominator
    for _ in range(max_digits):
        remainder *= 10
        result += str(remainder // denominator)
        remainder %= denominator
        if remainder == 0:
            break
    return result

Solution:
# find the 100th digit.
numerator = 13
denominator = 90
n = 100

# Call the function and print the result.
result = nth_digit_of_decimal_representation(numerator, denominator, n)
print(result)

----------------------------------------------------
Question:
The square root of $x$ is greater than 3 and less than 4. How many integer values of $x$ satisfy this condition?

Functions:
def count_integers_in_range(lower_bound, upper_bound, exclude_zero=True):
    """
    Counts the number of integers within a given range.

    Parameters:
    - lower_bound (int): The lower bound of the range.
    - upper_bound (int): The upper bound of the range.
    - exclude_zero (bool): Whether to exclude 0 from the count. Default is True.

    Returns:
    - int: The number of integers within the range.
    """
    count = upper_bound - lower_bound + 1
    if exclude_zero and lower_bound <= 0 and upper_bound >= 0:
        count -= 1
    return count

def find_integer_square_less_than_double():
    """
    Finds the only integer whose square is less than its double.

    Returns:
    - int: The integer that satisfies the condition.

    Method:
    - Iterate through integers starting from 1, and check if the square of the integer is less than its double.
    - If the condition is satisfied, return the integer.
    - If the condition is not satisfied for any integer up to a certain limit, return None.
    """
    limit = 100
    for x in range(1, limit):
        if x**2 < 2*x:
            return x
    return None

def solve_square_root_equation(a, b, c):
    """
    Solves a square root equation of the form sqrt(ax - b) = c.

    Parameters:
    - a (float): Coefficient of x inside the square root.
    - b (float): Constant term inside the square root.
    - c (float): Constant term on the right side of the equation.

    Returns:
    - float: The value of x that satisfies the equation.

    Formula:
    - x = (c^2 + b) / a
    """
    return (c**2 + b) / a

Solution:
# We need to find the integer values of x for which sqrt(x) > 3 and sqrt(x) < 4. To this end, we can count the number of integers in the range (9, 16) directly.
result = 0
for x in range(10, 16):
    if 9 < x < 16:
        result += 1
print(result)
\end{lstlisting}

\subsection{Evaluation without Toolsets}
\label{appendix:evaluation_without_toolsets}
\begin{lstlisting}[caption={Prompt used for evaluation (setting without toolsets)}, style=prompt]
Read the following questions and answer them. For each question, you are required to write a Python program to solve it.
Please note that we provide you several functions for each question. You could (optionally, not required) call the functions to help you to solve the question if necessary.
Note that the last line of your program should be a 'print' command to print the final answer

----------------------------------------------------
Question:
What is the 100th digit to the right of the decimal point in the decimal representation of $\\frac{13}{90}$?

Solution:
from decimal import Decimal, getcontext

# Set the precision to 101 (100 digits after decimal + 1 digit before decimal)
getcontext().prec = 101

# Calculate the decimal representation of 13/90
dec = Decimal(13) / Decimal(90)

# Convert the decimal to a string
dec_str = str(dec)

# Get the 100th digit to the right of the decimal point
digit_100th = dec_str[101]

print(digit_100th)

----------------------------------------------------
Question:
The square root of $x$ is greater than 3 and less than 4. How many integer values of $x$ satisfy this condition?

Solution:
# We need to find the integer values of x for which sqrt(x) > 3 and sqrt(x) < 4. To this end, we can count the number of integers in the range (9, 16) directly.
result = 0
for x in range(10, 16):
    if 9 < x < 16:
        result += 1
print(result)
\end{lstlisting}